\documentclass[twoside,leqno,twocolumn]{article}

\usepackage[letterpaper]{geometry}

\usepackage{ltexpprt}
\usepackage{hyperref}       % hyperlinks

\usepackage{url}            % simple URL typesetting
\usepackage{amsfonts}       % blackboard math symbols
\usepackage{nicefrac}       % compact symbols for 1/2, etc.
\usepackage{microtype}      % microtypography
\usepackage{xcolor}         % colors
\usepackage{minitoc}
\usepackage{wrapfig}
\usepackage{enumitem}
\usepackage{marvosym} 

\usepackage{algorithm}
\usepackage{minitoc}
\usepackage{multirow}
\usepackage{bm}
\usepackage{mathtools}
\usepackage{amsmath}
\usepackage{soul}
\usepackage{lipsum}
\usepackage{graphicx}
\usepackage{pythonhighlight}

\newtheorem{assumption}{Assumption}

\newcommand{\our}[0]{\textsc{DyFormer}~}
\newcommand*{\oure}{\textsc{DyFormer}}

\begin{document}

\newcommand\relatedversion{}
\renewcommand\relatedversion{\thanks{The full version of the paper can be accessed at \protect\url{https://arxiv.org/abs/2111.10447}}} % Replace URL with link to full paper or comment out this line

\title{\our: A Scalable Dynamic Graph Transformer with Provable Benefits on Generalization Ability}
% \author{Corey Gray\thanks{Society for Industrial and Applied Mathematics.}
% \and Tricia Manning\thanks{Society for Industrial and Applied Mathematics.}}
\author{
Weilin Cong\thanks{Penn State University.}~\thanks{Email: \url{wxc272@psu.edu}} \and Yanhong Wu\thanks{Meta AI.} \and Yuandong Tian$^\dagger$\and Mengting Gu$^\dagger$ \and Yinglong Xia$^\dagger$ \and  Chun-cheng Jason Chen$^\dagger$ \and Mehrdad Mahdavi$^*$ 
}
\date{}

\maketitle

% Copyright Statement
% When submitting your final paper to a SIAM proceedings, it is requested that you include
% the appropriate copyright in the footer of the paper.  The copyright added should be
% consistent with the copyright selected on the copyright form submitted with the paper.
% Please note that "20XX" should be changed to the year of the meeting.

% Default Copyright Statement
\fancyfoot[R]{\scriptsize{Copyright \textcopyright\ 2023 by SIAM\\
Unauthorized reproduction of this article is prohibited}}

% Depending on which copyright you agree to when you sign the copyright form, the copyright
% can be changed to one of the following after commenting out the default copyright statement
% above.

%\fancyfoot[R]{\scriptsize{Copyright \textcopyright\ 20XX\\
%Copyright for this paper is retained by authors}}

%\fancyfoot[R]{\scriptsize{Copyright \textcopyright\ 20XX\\
%Copyright retained by principal author's organization}}

%\pagenumbering{arabic}
%\setcounter{page}{1}%Leave this line commented out.

\begin{abstract} Transformers have achieved great success in several domains, including Natural Language Processing and Computer Vision. However, their application to real-world graphs is less explored, mainly due to its high computation cost and its poor generalizability caused by the lack of enough training data in the graph domain.
To fill in this gap, we propose a scalable Transformer-like dynamic graph learning method named \textbf{Dy}namic Graph Trans\textbf{former} (\oure) with \textit{spatial-temporal encoding} to effectively learn graph topology and capture implicit links. 
To achieve efficient and scalable training, we propose \textit{temporal-union graph} structure and its associated \textit{subgraph-based node sampling strategy}. 
To improve the generalization ability, we introduce two complementary \textit{self-supervised pre-training tasks} and show that jointly optimizing the two pre-training tasks results in a smaller Bayesian error rate via an information-theoretic analysis. 
Extensive experiments on the real-world datasets illustrate that \our achieves a consistent $1\%\sim3\%$ AUC gain (averaged over all time steps) compared with baselines on all benchmarks. 
\href{https://github.com/CongWeilin/DyFormer}{[\textcolor{blue}{Code}]}
\end{abstract}

\section{Introduction}
% \textcolor{red}{dynamic graph is important and has wide application.}

In recent years, graph representation learning has been recognized as a fundamental learning problem and has received much attention due to its widespread use in various domains, including social network analysis~\cite{kipf2016semi}, traffic prediction~\cite{rahimi2018semi}, knowledge graphs~\cite{wang2019knowledge}, drug discovery~\cite{do2019graph}, and recommender systems~\cite{berg2017graph}.
Most existing graph representation learning works focus on static graphs.
However, real-world graphs are intrinsically dynamic where nodes and edges can appear and disappear over time.
For example, the Facebook social network can be considered as a giant dynamic graph, where a new node is created when a user registers the account, and an edge between two nodes is created when a user connects to another one as a friend.
The dynamic nature of real-world graphs motivates graph learning methods that can model temporal evolutionary patterns and predict node properties or future links.

Although dynamic graph is important and has wide application, solving real-world dynamic graph learning problems is more challenging than traditional static graph learning problem due to the following reasons: \textit{\textbf{(1) missing or spurious links in dynamic graph}}: Real-world static graphs are potentially affected by missing/spurious links, applying Graph Neural Networks (GNNs) on real-world graphs could result in ineffective message aggregation over unrelated neighbors from missing/spurious connections. 
The issue is more severe on dynamic graphs because GNNs cannot distinguish whether it is missing/spurious links or is the temporal evolutionary pattern of the dynamic graph, which could potentially lead to poor generalization. 
Although several attempts~\cite{sankar2018dynamic,pareja2020evolvegcn,goyal2018dyngem,xu2020inductive} have been made to generalize the static graph algorithm to dynamic graphs by first learning node representations on each static graph snapshot then aggregating these representations from the temporal dimension, these methods still suffer from the aforementioned missing/spurious links issue. Furthermore, aggregating information on the temporal dimension could further carry such error over time, which can significantly affect downstream task accuracy; \textit{\textbf{(2) scalability issue at the temporal dimension}}: Unlike a fixed-size static graph, the size of a dynamic graph can increase over time. The complexity of most static graph GNNs is dependent on graph sizes, which makes these algorithms not scalable on large graphs~\cite{graphsaint, graphzoom}. Dynamic graphs introduce an additional level of complexity dependency on the number of time steps, which makes the computation issue more severe. Motivated by the importance and wide applications of dynamic graphs, we propose \our to solve the aforementioned challenges.

% \textcolor{red}{our proposal that solved the aformentioned challenges}

To overcome \textbf{\textit{the missing or spurious links issue}}, \textsc{DyFormer} leverages the Transformer~\cite{vaswani2017attention} as the backbone to model all pair-wise node relations using the fully-connected self-attention mechanism. By doing so, \our can model the relation between node pairs that have no links existed in the original graph,
% capture implicit edge connections,
thus becoming robust to graphs with missing and spurious links.
Meanwhile, to fully take advantage of the existing spatial and temporal information from the given dynamic graphs, we generalize the positional encoding to the graph domain using spatial-temporal encoding (Section~\ref{sec:spatial_temporal_encoding}) by injecting both spatial and temporal graph evolutionary information as inductive biases into \oure, which can 
help our model better utilize the existing graph structure and 
learn a graph's evolutionary patterns over time.
Furthermore, to alleviate the potential poor generalization ability caused by missing/spurious links, we introduce two complementary dynamic graph pre-training tasks that help \our present a better performance on the downstream tasks (Section~\ref{section:pretrain}) and a provable benefit on generalization ability using information theory.
To improve the \textbf{\textit{scalability issue}}, we propose to make the complexity independent on both the graph size and the number of time-steps. To achieve this, we first introduce the \textit{temporal-union graph} structure that aggregates graph information from multiple time-steps into a unified meta-graph (Section~\ref{sec:preprocessing}). Then, we develop a two-tower architecture (Section~\ref{section:model_arc}) with a novel subgraph-based node sampling strategy (Section~\ref{sec:mini_batch}) to model a subset of nodes with their contextual information.
These approaches improve~\oure's training efficiency and scalability from temporal and spatial perspectives.

To this end, we summarize our contributions as follows:
($1$) a two-tower Transformer-based method named~\our with the spatial-temporal encoding that can capture implicit edge connections in addition to the input graph topology;
($2$) two complementary pre-training tasks to improve generalization ability and robustness to missing/spurious links, which are proven beneficial using information theory;
($3$) a \textit{temporal-union graph} data structure that efficiently summarizes the spatial-temporal information of dynamic graphs and a novel sampling strategy that makes \our have complexity independent on graph size and the number of time steps;
and ($4$) a comprehensive evaluation on real-world datasets with ablation studies to validate the effectiveness of \oure.

\section{Preliminaries and related works}\label{sec:related}
We first define dynamic graphs, then review related works on dynamic graph and graph Transformers.

\paragraph{Dynamic graph definition.}
The nodes and edges in a dynamic graph may appear and disappear over time.
We consider a dynamic graph as a sequence of static graph snapshots with a temporal order $\smash{\mathbb{G}:= \{ \mathcal{G}_1,\ldots, \mathcal{G}_T\}}$, 
where the $t$-th snapshot graph $\mathcal{G}_t(\mathcal{V}, \mathcal{E}_t)$ is an undirected graph with a shared node set $\mathcal{V}$ of all time steps and an edge set $\mathcal{E}_t$. We also denote its adjacency matrix as $\mathbf{A}_t$.
Our goal is to learn the node representation at each time-step $t$, which can be used for any specific downstream task such as link prediction or node classification.
Please notice that our setting is the same  as the dynamic graph learning setting in~\cite{sankar2018dynamic,pareja2020evolvegcn}, where 
dynamic graph is defined as a set of temporal ordered snapshot graphs, in which 
the shared node set $\mathcal{V}$ are updated when new snapshot graph arrives.

\paragraph{Dynamic graph learning.}
Previous dynamic graph representation learning methods usually extend static graph algorithms by further taking the temporal information into consideration.
They can mainly be classified into three categories:
% as \textit{embedding-based method}, \textit{recurrent-based method}, and \textit{attention-based method}.
($1$) \textit{smoothness-based methods} learn a graph autoencoder to generate node embeddings on each graph snapshot and ensure the temporal smoothness of the node embeddings across consecutive time-steps.
For example, \textsc{DyGEM}~\cite{goyal2018dyngem} uses the learned embeddings from the previous time-step to initialize the embeddings in the next time-step.
\textsc{DynAERNN} applies RNN to smooth node embeddings at different time-steps;
($2$) \textit{Recurrent-based methods} capture the temporal dependency using RNN.
For example, \textsc{GCRN}~\cite{seo2018structured} first computes node embeddings on each snapshot using GCN~\cite{defferrard2016convolutional}, then feeds the node embeddings into an RNN to learn their temporal dependency.
\textsc{EvolveGCN}~\cite{pareja2020evolvegcn} uses RNN to estimate the GCN weight parameters at different time-steps; 
($3$) \textit{Attention-based methods} use self-attention mechanism for both spatial and temporal message aggregation.
For example, \textsc{DySAT}~\cite{sankar2018dynamic}  propose to use the self-attention mechanism for both temporal and spatial information aggregation.
\textsc{TGAT}~\cite{xu2020inductive} encodes the temporal information into the node feature, then applies self-attention on the temporal augmented node features.
However, \textit{smoothness-based methods} heavily rely on temporal smoothness and are inadequate when nodes exhibit vastly different evolutionary behaviors,
\textit{recurrent-based methods} scale poorly when the number of time-steps increases due to RNN's recurrent nature,
\textit{attention-based methods} only consider the self-attention on existing edges and are sensitive to missing/spurious links in graphs.
In contrast,~\our~leverages Transformer to capture the spatial-temporal dependency between all nodes pairs, does not over-rely on the given graph structures, and is less sensitive to missing/spurious links. 

\paragraph{Graph Transformers.}
Recently, several attempts have been made to leverage Transformer for graph representation learning.
For example, 
\textsc{Graphormer}~\cite{ying2021transformers} and \textsc{GraphTransformer}~\cite{dwivedi2020generalization} use scaled dot-product attention~\cite{vaswani2017attention} for message aggregation and 
generalizes the idea of positional encoding to graph domains.
\textsc{GraphBert}~\cite{zhang2020graph} first samples an egocentric network for each node, then orders all nodes into a sequence based on node importance, and feed into the Transformer.
However, \textsc{Graphormer}~\cite{ying2021transformers} is only feasible to small molecule graphs and cannot scale to large graphs due to the significant computation cost of full attention; \textsc{GraphTransformer}~\cite{dwivedi2020generalization} only considers the first-hop neighbor aggregation, which makes it sensitive to noisy graphs;
\textsc{GraphBert}~\cite{zhang2020graph} does not leverage the graph topology and can perform poorly when graph topology is important.
In contrast, \our encodes the input graph structures as an inductive bias to guide the full-attention optimization, which balances the trade-offs between noisy input robustness and efficiently learning an underlying graph structure. 

\section{Method}

\begin{figure*}[t]
    \centering
    \includegraphics[width=0.8\textwidth]{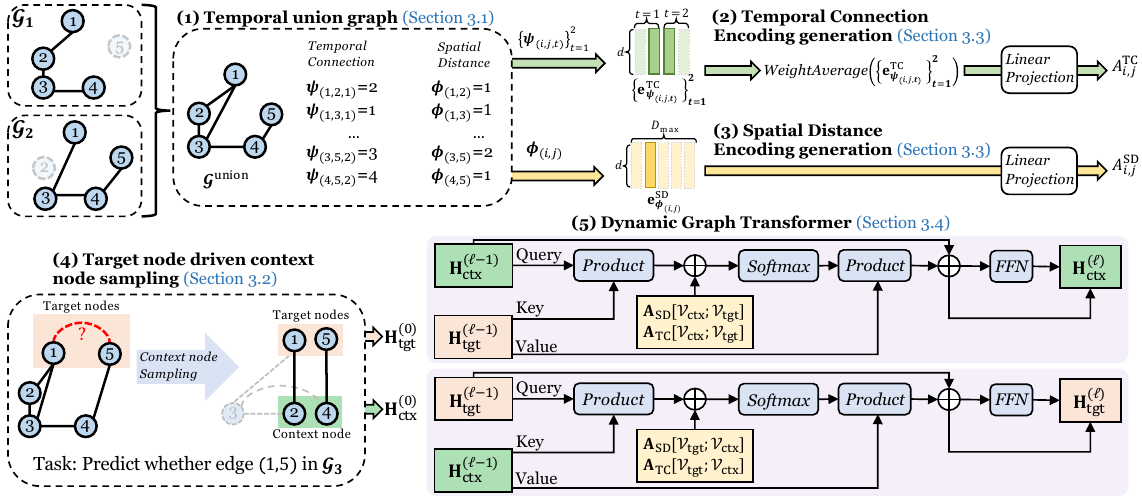}
    \vspace{-5mm}
    \caption{Overview of using \our for link prediction. Given snapshot graphs $\{\mathcal{G}_1, \mathcal{G}_2\}$ as input, \textbf{(1)} we first generate the temporal union graph with the considered max shortest path distance $D_{\max}=5$, and its associated \textbf{(2)} temporal connection encoding and \textbf{(3)} spatial distance encoding. Then, the encodings are mapped into $A_{i,j}^\text{TC},A_{i,j}^\text{SD}$ for each node pairs $(i,j)$ using a fully connected layer. To predict whether an edge exists in $\mathcal{G}_3$, we first \textbf{(4)} sample target and context nodes, then apply \textbf{(5)} \our to encode target nodes and context nodes separately.}
    \label{fig:pipeline}
\end{figure*}

In this section, we first introduce the temporal union-graph (Section~\ref{sec:preprocessing}) and our sampling strategy (Section~\ref{sec:mini_batch}) that can reduce the overall complexity from the temporal and spatial perspectives.
Then, we introduce spatial-temporal encoding technique (Section~\ref{sec:spatial_temporal_encoding}), describe the two-tower transformer architecture design, and explain how to integrate the spatial-temporal encoding to \our (Section~\ref{section:model_arc}).
Figure~\ref{fig:pipeline} illustrates the overall \our design.

%%%%%%%%%%%%%%%%%%%%%%%%%%%%%%%%%%%%%%%%%%%%%%%%%%%%%%%%%%
%%%%%%%%%%%%%%%%%%%%%%%%%%%%%%%%%%%%%%%%%%%%%%%%%%%%%%%%%%
%%%%%%%%%%%%%%%%%%%%%%%%%%%%%%%%%%%%%%%%%%%%%%%%%%%%%%%%%%

\subsection{Temporal-union graph generation}
\label{sec:preprocessing}

One major challenge of applying Transformers on graph representation learning is its significant computation and memory overhead.
In Transformers, the computation cost of self-attention is $\mathcal{O}(|\mathcal{E}|\, d)$ and its memory cost is $\mathcal{O}(|\mathcal{E}|+|\mathcal{V}|\,d)$. 
When using full attention, the computation graph is fully connected with $|\mathcal{E}|=|\mathcal{V}|^2$, where the overall complexity is quadratic in the graph size.
Although \textsc{Linformer}~\cite{wang2020linformer} and \textsc{BigBird}~\cite{zaheer2020big} reduce its complexity to $\mathcal{O}(|\mathcal{V}| d)$ by using sparse self-attention, it is still computationally prohibitive as the size of real-world graph are easily sized to the billion level.
On dynamic graphs, this problem can be even more severe if one naively extends the static graph algorithm to a dynamic graph, e.g., first extracting the spatial information of each snapshot graph separately, then jointly reasoning the temporal information on all snapshot graphs~\cite{sankar2018dynamic,pareja2020evolvegcn}.
By doing so, the overall complexity grows linearly with the number of time-steps $T$, i.e., with $\mathcal{O}(|\mathcal{V}|^2 Td)$ computation and  $\mathcal{O}(|\mathcal{V}|^2 T+|\mathcal{V}|Td)$ memory cost. 
To reduce the dependency of the overall complexity on the number of time-steps,
we propose to first aggregate dynamic graphs $\mathbb{G}=\{ \mathcal{G}_1,\ldots, \mathcal{G}_T\}$ into a \textit{temporal-union graph} $\mathcal{G}^\text{union}(\mathcal{V}, \mathcal{E}^\prime)$ then employ \our on the generated temporal-union graph, where $\mathcal{E}^\prime=\textit{Unique}\{(i,j): (i,j)\in\mathcal{E}_t,~t\in[T]\}$ is the set of all possible unique edges in $\mathbb{G}$.
As a result, the overall complexity of \our does not grow with the number of time-steps.
Details on how to leverage spatial-temporal encoding to recover the temporal information of edges are described in Section~\ref{sec:spatial_temporal_encoding}.

%%%%%%%%%%%%%%%%%%%%%%%%%%%%%%%%%%%%%%%%%%%%%%%%%%%%%%%%%%
%%%%%%%%%%%%%%%%%%%%%%%%%%%%%%%%%%%%%%%%%%%%%%%%%%%%%%%%%%
%%%%%%%%%%%%%%%%%%%%%%%%%%%%%%%%%%%%%%%%%%%%%%%%%%%%%%%%%%
\subsection{Target node driven context node sampling} \label{sec:mini_batch}

Although the temporal-union graph can alleviate the computation burden from the temporal dimension, due to the overall quadratic  complexity of self-attention with respect to the input graph size, scaling the training of Transformer to real-world graphs is still non-trivial.
Therefore, a properly designed sampling strategy that makes the overall complexity independent with graph sizes is necessary.
Our goal is to design a sub-graph sampling strategy that ensures a fixed number of well-connected nodes and a lower computational complexity.
To this end, we propose to first sample a subset of nodes that we are interested in as \textit{target nodes}, then sample their common neighbors as \textit{context nodes}. 

Let \textit{target nodes} $\mathcal{V}_\text{tgt} \subseteq \mathcal{V}$ be the nodes that we are interested in and want to compute its node representation. For example, for the link prediction task,
$\mathcal{V}_\text{tgt}$ are the set of nodes that we aim to predict whether they are connected.
Then, the \textit{context nodes} $\mathcal{V}_\text{ctx} \subseteq \{\mathcal{N}(i)~|~\forall i \in \mathcal{V}_\text{tgt}\}$ are sampled as the common neighbors of the target nodes.
Notice that since context nodes $\mathcal{V}_\text{ctx}$ are sampled as the common neighbors of the target nodes, they can provide local structure information for nodes in the target node set. Besides, since two different nodes in the target node set can be far apart with a disconnected neighborhood, the neighborhood of two nodes can provide an approximation of the global view of the full graph. 
During sampling, to control the randomness involved in the sampling process, $\mathcal{V}_\text{ctx}$ are chosen as the subset of nodes with the top-$K$ joint Personalized PageRank (PPR) score~\cite{andersen2006local} to nodes in $\mathcal{V}_\text{tgt}$, where PPR score is a node proximity measure that captures the importance of two nodes in the graph. 
More specifically, our joint PPR sampler proceeds as follows: First, we compute the approximated PPR vector $\bm{\pi}(i)\in\mathbb{R}^{N}$ for all node $i\in \mathcal{V}_\text{tgt}$, where the $j$-th element in $\bm{\pi}(i)$ can be interpreted as the probability of a random walk to
start at node $i$ and end at node $j$. We then compute the approximated joint PPR vector $\hat{\bm{\pi}}(\mathcal{V}_\text{tgt}) = \sum_{i\in\mathcal{V}_\text{tgt}} \bm{\pi}(i) \in\mathbb{R}^{N} $. Finally, we select $K$ context nodes where each node $j\in\mathcal{V}_\text{ctx}$ has the top-$K$ joint PPR score in $\hat{\bm{\pi}}(\mathcal{V}_\text{tgt})$. In practice, 
the context node size $K$ is the same as the target node size $|\mathcal{V}_\text{tgt}|$.

%%%%%%%%%%%%%%%%%%%%%%%%%%%%%%%%%%%%%%%%%%%%%%%%%%%%%%%%%%
%%%%%%%%%%%%%%%%%%%%%%%%%%%%%%%%%%%%%%%%%%%%%%%%%%%%%%%%%%
%%%%%%%%%%%%%%%%%%%%%%%%%%%%%%%%%%%%%%%%%%%%%%%%%%%%%%%%%%

\subsection{Spatial-temporal encoding}\label{sec:spatial_temporal_encoding}
Given a temporal-union graph, our next step is to translate the spatial-temporal information from snapshot graphs to the temporal-union graph $\mathcal{G}_\text{union}$, which can be recognized and leveraged by Transformers.
Most classical GNNs either over-rely on the given graph structure by only considering the first- or higher-order neighbors for feature aggregation~\cite{ying2021transformers} (which could make the model fail to capture the inter-relation between nodes that are not connected in the labeled graph) or directly learn graph adjacency without using the given graph structure~\cite{devlin2018bert} (which makes the optimization problem challenging because the model has to iteratively learn model parameters and estimate the graph structure).
To avoid the above two extremes, we present two simple but effective encoding designs, i.e., \textit{temporal connection encoding} and \textit{spatial distance encoding}, and introduce how to integrate them into \oure. 

\paragraph{Temporal connection encoding.}
Temporal connection (TC) encoding is designed to inform \our if an edge $(i,j)$ exists in the $t$-th snapshot graph.
We denote $\mathbf{E}^\text{TC} = [\mathbf{e}_{2t-1}^\text{TC}, \mathbf{e}_{2t}^\text{TC}]_{t=1}^T \in \mathbb{R}^{2T\times d}$ as the temporal connection encoding lookup-table where $d$ represents the hidden dimension size, which is indexed by a function $\psi(i,j,t)$ indicating whether an edge $(i,j)$ exists at time-step $t$.
More specifically, we have $\psi(i,j,t)=2t$ if $(i,j) \in \mathcal{G}_t$, $\psi(i,j,t)=2t-1$ if $(i,j) \not\in \mathcal{G}_t$ and use this value as an index to extract the corresponding temporal connection embedding from the look-up table for next-step processing.
Note that during pre-training or the training on first few time-steps, we need to mask-out certain time-steps to avoid leaking information related to the predicted items (e.g., the temporal reconstruction task in Section.~\ref{section:pretrain}).
In these cases, we set $\psi(i,j,t')=\O$ where $t'$ denotes the time-step we mask-out, and skip the embedding extraction at time $t'$. 

%%%%%%%%%%%%%%%%%%%%%%%%%%%%%%%%%%%
\paragraph{Spatial distance encoding.}
Spatial distance (SD) encoding is designed to provide \our a global view of the graph structure.
The success of Transformer is largely attributed to its global receptive field due to its full attention, i.e., each token in the sequence can attend independently to other tokens and process its representations.
Computing full attention requires the model to explicitly capturing the positions dependency between tokens, which can be achieved by either assigning each position an absolute positional encoding or encode the relative distance using relative positional encoding.
However, for graphs, the design of unique node positions is not mandatory because a graph is not changed by the permutation of its nodes.
To encode the global structural information of a graph in the model, inspired by~\cite{ying2021transformers}, we adopt a spatial distance encoding that measures the relative spatial relationship between any two nodes in the graph, which is a generalization of the classical Transformer's positional encoding to the graph domain.
Let $D_{\max}$ be the maximum shortest path distance (SPD) we considered, where $D_{\max}$ is a hyper-parameter that can be smaller than the graph diameter. Specifically, given any node $i$ and node $j$, we define $\phi(i,j) = \min\{\text{SPD}(i,j),~D_{\max}\}$ as the SPD between the two nodes if  $\text{SPD}(i,j) < D_{\max}$ and otherwise as $D_{\max}$.  Let $\mathbf{E}^\text{SD} = [ \mathbf{e}^\text{SD}_{1},\ldots,\mathbf{e}^\text{SD}_{D_{\max}}] \in\mathbb{R}^{D_{\max} \times d}$ as the spatial distance lookup-table which is indexed by the $\phi(i,j)$, where $\phi(i,j)$ is used to select the spatial distance encoding $\smash{\mathbf{e}_{\phi(i,j)}^\text{SD}}$ that provides the spatial distance information of two nodes. 

%%%%%%%%%%%%%%%%%%%%%%%%%%%%%%%%%%%%%%%%%%%%%%%%%%%%%%%%%%
%%%%%%%%%%%%%%%%%%%%%%%%%%%%%%%%%%%%%%%%%%%%%%%%%%%%%%%%%%
%%%%%%%%%%%%%%%%%%%%%%%%%%%%%%%%%%%%%%%%%%%%%%%%%%%%%%%%%%
\paragraph{Integrate spatial-temporal encoding.}
We integrate temporal connection encoding and spatial distance encoding by 
% first projecting them as scalar using the fully connected layer, then treat the projected scalar 
projecting them as a bias term in the self-attention module.
Specifically, to integrate the \textit{spatial-temporal encoding} of node pair $(i,j)$ to \oure, 
we first gather all its associated temporal connection encodings on different time-steps as $\{\mathbf{e}_{\phi(i,j,t)}^\text{TC}\}_{t=1}^T$. Then, we apply weight average on all encodings over the temporal axis and projected the temporal averaged encoding as a scalar by $A_{i,j}^\text{TC} = \textit{Linear} \big( \textit{WeightAverage}(\{\mathbf{e}_{\phi(i,j,t)}^\text{TC}\}_{t=1}^T) \big) \in \mathbb{R}$, where the aggregation weight is learned during training.
Similarly, to integrate the \textit{spatial distance encoding},
we project the spatial distance encoding of node  pair $(i,j)$ as a scalar by $ A_{i,j}^\text{SD} = \textit{Linear}(\mathbf{e}_{\phi(i,j)}^\text{SD}) \in \mathbb{R}$.
Then, $A_{i,j}^\text{TC}$ and $A_{i,j}^\text{SD}$ are used as the bias term to the self-attention, which we describe in detail in Section~\ref{section:model_arc}.

%%%%%%%%%%%%%%%%%%%%%%%%%%%%%%%%%%%%%%%%%%%%%%%%%%%%%%%%%%
%%%%%%%%%%%%%%%%%%%%%%%%%%%%%%%%%%%%%%%%%%%%%%%%%%%%%%%%%%
%%%%%%%%%%%%%%%%%%%%%%%%%%%%%%%%%%%%%%%%%%%%%%%%%%%%%%%%%%

\subsection{Graph Transformer architecture} \label{section:model_arc}

% Based on the mini-batch sampling strategy and spatial-temporal encoding strategy mentioned above,
% Now we are ready to introduce \oure. 
As shown in Figure~\ref{fig:pipeline}, each layer in \our consists of two towers (i.e., the target node tower and the context node tower) to encode the target nodes and the context nodes separately.
The same set of parameters are shared between two towers.
The two-tower structure is motivated by the fact that nodes within each group are sampled independently but there exist neighborhood relationships between inter-group nodes. Only attending inter-group nodes help \our better capture this context information without fusing representations from irrelevant nodes.
% In addition, the two-tower structure can help us alleviate the computation burden, and allow us to use a larger mini-batch with a more stabilized stochastic gradient.
In the following, we provide details on the context node tower and the detailed formulation for the target node tower can be obtained by switching ``ctx'' with ``tgt''.

\begin{itemize} [noitemsep,topsep=0pt,leftmargin=3mm ]
    \item First, we compute the self-attentions to aggregate information from target nodes to context nodes (denote as ``$\text{ctx}$'') and from context nodes to target nodes (denote as ``$\text{tgt}$'').
    Let define
    $\smash{\mathbf{H}_\text{ctx}^{(\ell)}} \in \mathbb{R}^{|\mathcal{V}_\text{ctx}| \times d}$ as the $\ell$-th layer output of the context-node tower and $\smash{\mathbf{H}_\text{tgt}^{(\ell)}} \in \mathbb{R}^{|\mathcal{V}_\text{tgt}| \times d}$ as the $\ell$-th layer output of the target-node tower.
    Then, the $\ell$th layer self-attention is
    \vspace{-2mm}
    \begin{equation*}
        \mathbf{A}^{(\ell)}_{\text{ctx}} = \frac{(\texttt{LN}( \mathbf{H}_\text{ctx}^{(\ell-1)} )\mathbf{W}_Q^{(\ell)} ) (\texttt{LN}(\mathbf{H}_\text{tgt}^{(\ell-1)}) \mathbf{W}_K^{(\ell)} )^\top  }{\sqrt{d}}
        ,~
        % \mathbf{A}^{(\ell)}_{\text{tgt}} = \frac{(\texttt{LN}(\mathbf{H}_\text{tgt}^{(\ell-1)} )\mathbf{W}_Q^{(\ell)}) (\texttt{LN}(\mathbf{H}_\text{ctx}^{(\ell-1)}) \mathbf{W}_K^{(\ell)})^\top  }{\sqrt{d}},
        \vspace{-2mm}
    \end{equation*}
    where $\texttt{LN}(\mathbf{H})$ stands for applying layer normalization on $\mathbf{H}$ and
    $\smash{\mathbf{W}_Q^{(\ell)}, \mathbf{W}_K^{(\ell)}}$ are weight matrices.
    \item Then, we 
    integrate spatial-temporal encoding as a bias term to self-attention as 
    \vspace{-2mm}
    \begin{equation*}
        \mathbf{P}^{(\ell)}_{\text{ctx}}  = \mathbf{A}^{(\ell)}_{\text{ctx}} + \mathbf{A}_\text{TC} [\mathcal{V}_\text{ctx}; \mathcal{V}_\text{tgt} ] + \mathbf{A}_\text{SD} [\mathcal{V}_\text{ctx}; \mathcal{V}_\text{tgt} ],~~
        % \mathbf{P}^{(\ell)}_{\text{tgt}} = \mathbf{A}^{(\ell)}_{\text{tgt}} + \mathbf{A}_\text{TC} [\mathcal{V}_\text{tgt}; \mathcal{V}_\text{ctx} ] + \mathbf{A}_\text{SD} [ \mathcal{V}_\text{tgt}; \mathcal{V}_\text{ctx} ],
        \vspace{-2mm}
    \end{equation*}
    where $\mathbf{A}_\text{TC} [\mathcal{V}_A; \mathcal{V}_B ],~\mathbf{A}_\text{SD} [\mathcal{V}_A; \mathcal{V}_B ]$ denote the matrix form of the projected temporal connection and spatial distance self-attention bias with row and column indexed by $\mathcal{V}_A$ and $\mathcal{V}_B$.\footnote{Given a matrix $\mathbf{A} \in \mathbb{R}^{m \times n}$, the element at the $i$-th row and $j$-th column is denoted as $A_{i,j}$, the submatrix formed from row $\mathcal{I}_\text{row}=\{a_1,\ldots, a_r\}$ and columns $\mathcal{I}_\text{col} = \{b_1, \ldots, b_s\}$ is denoted as $\mathbf{A} \left[\mathcal{I}_\text{row}; \mathcal{I}_\text{col}\right]$.}
    \item After that, we use the normalized $\mathbf{P}^{(\ell)}_{\text{ctx}}$ and $\mathbf{P}^{(\ell)}_{\text{tgt}}$ to propagate information between two towers, i.e.,
    \vspace{-2mm}
    \begin{equation*}
        \mathbf{Z}_\text{ctx}^{(\ell)} = \textit{Softmax} (\mathbf{P}^{(\ell)}_{\text{ctx}} )\texttt{LN}(\mathbf{H}_\text{tgt}^{(\ell-1)}) \mathbf{W}_V^{(\ell)}  + \mathbf{H}_\text{ctx}^{(\ell-1)},~~
        % \mathbf{Z}_\text{tgt}^{(\ell)}  = \textit{Softmax}(\mathbf{P}^{(\ell)}_{\text{tgt}}) \texttt{LN}(\mathbf{H}_\text{ctx}^{(\ell-1)}) \mathbf{W}_V^{(\ell)} + \mathbf{H}_\text{tgt}^{(\ell-1)}.
        \vspace{-2mm}
    \end{equation*}
    \item Finally, a residual connected feed-forward network is applied to the aggregated message to produce the final output $\mathbf{H}_\text{ctx}^{(\ell)} = \textit{FFN}(\texttt{LN}(\mathbf{Z}_\text{ctx}^{(\ell)} )) + \mathbf{Z}_\text{ctx}^{(\ell)}$ where $\textit{FFN}(\cdot)$ denotes the multi-layer feed-forward network. The final layer output of the target node tower $\smash{\mathbf{H}_\text{tgt}^{(L)}}$ will be used to compute the loss defined in Section~\ref{section:learning_pretrain_and_finetune}.
\end{itemize}

\section{\our training} \label{section:learning_pretrain_and_finetune}

Transformers usually require a significant amount of supervised data to guarantee their generalization ability on unseen data. 
However, existing dynamic graph datasets are relatively small and may not be sufficient to train a powerful Transformer.
To overcome this challenge, we propose to first pre-train \our with two complementary self-supervised objective functions (in Section~\ref{section:pretrain}).
Then, we fine-tune \our using the supervised objective function (in Section~\ref{section:finetune}). 
% Notice that suppose our goal is to predict links at time $T+1$ using all snapshot graph until time $T$, then both pre-training and fine-tuning only have access to snapshot graph until time $T$.
Notice that the same set of snapshot graphs but different objective functions are used for pre-training and fine-tuning.
Finally, via an information-theoretic analysis, we show that the representation can enjoy a better generalization ability on downstream tasks by optimizing our pre-training losses (in Section~\ref{section:information_theory}).

\subsection{Pre-training} \label{section:pretrain}

\begin{figure}[t]
    \centering
    % \vspace{-10pt}
    \includegraphics[width=0.51\textwidth]{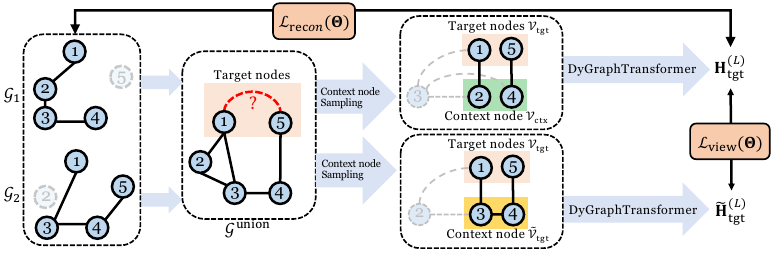}
    \vspace{-9mm}
    \caption{Pre-training: Given snapshot graphs $\{\mathcal{G}_1,\mathcal{G}_2\}$ as input, we first generate the temporal union graph. Then, we sample the target node $\mathcal{V}_\text{tgt}$ and two different set of context nodes $\smash{\mathcal{V}_\text{ctx},\widetilde{\mathcal{V}}_\text{ctx}}$. After that, we apply \our on $\{  \mathcal{V}_\text{tgt}, \mathcal{V}_\text{ctx} \}$ and $\{ \mathcal{V}_\text{tgt}, \smash{\widetilde{\mathcal{V}}_\text{ctx}} \}$ to output $\smash{\mathbf{H}_\text{tgt}^{(L)}}$ and $\smash{\widetilde{\mathbf{H}}_\text{tgt}^{(L)}}$. We optimize $\mathcal{L}_\text{view}(\mathbf{\Theta})$ by maximizing the similarity between $\smash{\mathbf{H}_\text{tgt}^{(L)}}$ and $\smash{\widetilde{\mathbf{H}}_\text{tgt}^{(L)}}$, and optimize $\mathcal{L}_\text{recon}(\mathbf{\Theta})$ by recovering snapshot graphs using $\smash{\mathbf{H}_\text{tgt}^{(L)}}$.}
    \label{fig:pretrain_los}
    \vspace{-2mm}
\end{figure}

We introduce a \textit{temporal reconstruction loss} $\mathcal{L}_\text{recon}(\bm{\Theta})$ and a \textit{multi-view contrastive loss} $\mathcal{L}_\text{view}(\bm{\Theta})$ as self-supervised object functions.
Then, our overall pre-taining loss is $\mathcal{L}_\text{pre-train}(\bm{\Theta})= \mathcal{L}_\text{recon}(\bm{\Theta}) + \gamma \mathcal{L}_\text{view}(\bm{\Theta})$, where $\gamma$ is a hyper-parameter that balances the importance of two pre-taining tasks as in Figure~\ref{fig:pretrain_los}.

\paragraph{Temporal reconstruction loss.} 
To ensure that the spatial-temporal encoding is effective and can inform \our the temporal dependency between multiple snapshot graphs, we introduce a temporal reconstruction loss as our first pre-training objective.
Our goal is to reconstruct the $t$-th graph snapshot $\mathcal{G}_t$'s structure using all graph snapshot $\mathbb{G}$.
% its past snapshot graphs $\mathbb{G}_{1}^{t-1} = \{\mathcal{G}_1,\ldots, \mathcal{G}_{t-1}\}$ and its future snapshot graphs $\mathbb{G}_{t+1}^{T}= \{\mathcal{G}_{t+1},\ldots, \mathcal{G}_T\}$.
Let $\smash{\mathbf{H}_\text{tgt}^{(L)}(t)}$ denote the target-node tower's final layer output computed on $\mathbb{G}$.
% \yh{We may simplify as sth like:} $\mathbb{G} \setminus \mathcal{G}_t$
% In the following, we ignore the subscript $\text{tgt}$ for brevity and denote it as $\mathbf{H}^{(L)}_{\text{tgt,}\bar{t}}$.
% contains the extracted spatial and temporal information of the dynamic graph.
To decode the graph structure of graph snapshot $\mathcal{G}_t$, we use a fully connected layer as the temporal structure decoder that takes $\smash{\mathbf{H}^{(L)}_\text{tgt}(t)}$ as input and output $\mathbf{E}(t) = \smash{\textit{Linear}(\mathbf{H}^{(L)}_\text{tgt}(t))} \in \mathbb{R}^{|\mathcal{V}_\text{tgt}|\times d }$
with $\mathbf{e}_i(t) \in \mathbb{R}^d$ denotes the $i$-th row of $\mathbf{E}(t)$.
Then, the temporal reconstruction loss is 
$\mathcal{L}_\text{recon}(\bm{\Theta}) = \smash{\sum_{t=1}^T \textit{LinkPredLoss}(\{\mathbf{e}_i(t)\}_{i\in\mathcal{V}_\text{tgt}}, \mathcal{V}_\text{tgt}, \mathcal{E}_t)}$,
% \begin{equation}\label{eq:recon_loss}
%     \mathcal{L}_\text{recon}(\bm{\Theta}) = \sum_{t=1}^T \textit{LinkPredLoss}(\mathbf{E}(t), \mathcal{V}_\text{tgt}, \mathcal{E}_t)
%     % \sum_{i\in\mathcal{V}_\text{tgt}} \left( - \sum_{j\in\mathcal{N}(i) \cap \mathcal{V}_\text{tgt}} \log \Big(\sigma\big( \langle \mathbf{e}_i(t), \mathbf{e}_j(t) \rangle \big) \Big) - \sum_{k \not\in \mathcal{N}(i) \wedge k\in \mathcal{V}_\text{tgt} } \log \Big(1-\sigma\big( \langle \mathbf{e}_i(t), \mathbf{e}_k(t) \rangle \big) \Big) \right),
% \end{equation}
where
$\sigma(\cdot)$ is Sigmoid function and 
$\textit{LinkPredLoss}(\{\mathbf{x_i}\}_{i\in\mathcal{S}}, \mathcal{S}, \mathcal{E}):=$ 
\vspace{-4mm}
\begin{equation*}
    \sum_{i,j\in\mathcal{S}} \Big( -\sum_{(i,j)\in\mathcal{E}} \log (\sigma(\mathbf{x}_i^\top \mathbf{x}_j)) - \sum_{(i,j)\not\in\mathcal{E}} \log (1-\sigma(\mathbf{x}_i^\top \mathbf{x}_j)) \Big).
    \vspace{-3mm}
\end{equation*}

% $\sum_{i,j\in\mathcal{S}} \Big( -\sum_{(i,j)\in\mathcal{E}} \log (\sigma(\mathbf{x}_i^\top \mathbf{x}_j)) - \sum_{(i,j)\not\in\mathcal{E}} \log (1-\sigma(\mathbf{x}_i^\top \mathbf{x}_j)) \Big).$

% \vspace{-2mm}
% \begin{equation*}
% % \vspace{-5pt}
% %  \textit{LinkPredLoss}(\{\mathbf{x_i}\}_{i\in\mathcal{S}}, \mathcal{S}, \mathcal{E}) \triangleq 
%  \sum_{i,j\in\mathcal{S}} \Big( -\sum_{(i,j)\in\mathcal{E}} \log (\sigma(\mathbf{x}_i^\top \mathbf{x}_j)) - \sum_{(i,j)\not\in\mathcal{E}} \log (1-\sigma(\mathbf{x}_i^\top \mathbf{x}_j)) \Big).
% %  \vspace{-5pt}
% \vspace{-2mm}
% \end{equation*} 

% and $\sigma(\cdot)$ is Sigmoid function.
% \yh{Did we define sigma?}
%%%%%%%%%%%%%%%%%%
\paragraph{Multi-view contrastive loss.~}
% \yh{qq: does this ~double our training cost? (double forward pass?)}
Recall that $\mathcal{V}_\text{ctx}$ is constructed by \textit{deterministically} selecting the common neighbors of $\mathcal{V}_\text{tgt}$ with the top-$K$ PPR score. Then, we introduce $\widetilde{\mathcal{V}}_\text{ctx}$ as the subset of the common neighbors of $\mathcal{V}_\text{tgt}$ \textit{randomly} sampled with sampling probability of each node proportional to its PPR score. 
Since a different set of context nodes are provided for the same set of target nodes, $\{\mathcal{V}_\text{tgt}, \widetilde{\mathcal{V}}_\text{ctx}\}$ provides an alternative view of $\{\mathcal{V}_\text{tgt}, \mathcal{V}_\text{ctx}\}$ when computing the representation for nodes in $\mathcal{V}_\text{tgt}$.
Notice that although the provided context nodes are different, since they have the same target nodes, it is natural to expect the calculated representation have high similarity.
We denote $\smash{\mathbf{H}_\text{tgt}^{(L)}}$ and $\smash{\widetilde{\mathbf{H}}_\text{tgt}^{(L)}}$ as the final layer model output that are computed on $\{\mathcal{V}_\text{tgt}, \mathcal{V}_\text{ctx}\}$ and $\{\mathcal{V}_\text{tgt}, \smash{\widetilde{\mathcal{V}}_\text{ctx}} \}$. 
To this end, we introduce our second self-supervised objective function as $\mathcal{L}_\text{view}(\bm{\Theta})  = \| \mathbf{H}_\text{tgt}^{(L)} - \textit{SG}(\widetilde{\mathbf{H}}_\text{tgt}^{(L)}) \|_\mathrm{F}^2 + \| \textit{SG}(\mathbf{H}_\text{tgt}^{(L)}) - \widetilde{\mathbf{H}}_\text{tgt}^{(L)} \|_\mathrm{F}^2$, where $\textit{SG}$ denotes stop gradient.
% Note that optimizing $\mathcal{L}_\text{view}(\bm{\Theta})$ alone without stopping gradient results in a degenerated solution~\cite{chen2021exploring,tian2021understanding}.

\subsection{Fine-tuning} \label{section:finetune}
To apply the pre-trained model on downstream tasks, we choose to fine-tune the pre-trained model with downstream task objective functions.
Here, we take link prediction as an example.
Our goal is to predict the existence of a link at time $T+1$ using information up to time $T$.
Let $\smash{\mathbf{H}_\text{tgt}^{(L)}(\{\mathcal{G}_j\}_{j=1}^t)}$ denote the final output of \our using snapshot graphs $\{\mathcal{G}_j\}_{j=1}^t$. Then, the link prediction loss is  $\mathcal{L}_\text{LinkPred}(\bm{\Theta}) = \sum_{t=1}^{T-1} \textit{LinkPredLoss}(\mathbf{H}_\text{tgt}^{(L)}(\{\mathcal{G}_j\}_{j=1}^t),\mathcal{V}_\text{tgt},\mathcal{E}_{t+1})$.

\subsection{Importance of pre-training} \label{section:information_theory}

In this section, we show that our pre-training objectives can improve the generalization error under mild assumptions and results in a better 
performance on downstream tasks.
Let $X$ denote the input random variable, $S$ as the self-supervised signal (also known as a different view of input $X$), and $Z_X = f(X), Z_S = f(S)$ as the representations that are generated by a deterministic mapping function $f$.
In our setting,  we have the sampled sub-graph of temporal-union graph $\mathcal{G}^\text{union}$ induced by node $\{\mathcal{V}_\text{tgt}, \mathcal{V}_\text{ctx}\}$ as input $X$, the sampled subgraph of $\mathcal{G}^\text{union}$ induced by node $\{\mathcal{V}_\text{tgt}, \smash{\widetilde{\mathcal{V}}_\text{ctx}} \}$ as self-supervised signal $S$, and \our as $f$ that computes the representation of $X,S$ by $Z_X = f(X), Z_S = f(S)$.
Besides, we introduce the task-relevant information as $Y$, which refers to the information that is required for downstream tasks.
For example, when the downstream task is link prediction, $Y$ can be the ground truth graph structure about which we want to reason.
Notice that in practice we have no access to $Y$ during pre-training and it is only introduced as the notation for analysis.
Furthermore, let $H(A)$ denote entropy, $H(A|B)$ denote conditional entropy, $I(A;B)$ denote mutual information, and $I(A;B|C)$ denote conditional mutual information. More details and preliminaries on information theory are deferred to Appendix~\ref{section:information_theory_self_supervised}.

In the following, we study the generalization error of the learned representation $Z_X$ under the binary classification setting. We choose \textit{Bayes error rate} (i.e., the lowest possible test error rate a binary classifier can achieve) as our evaluation metric, which can be formally defined as $P_e = 1 - \mathbb{E} [ \max_{y} \mathrm{P}(Y=y|Z_X)]$.
Before proceeding to our result, we make the following assumption on input $X$, self-supervised signal $S$, and task-relevant information $Y$.

\begin{assumption}\label{assumption:X_S_T_assump}
Assume task-relevant information is shared between input random variable $X$, self-supervised signal $S$, i.e., $I(X;Y|S) = 0$ and $I(S;Y|X)=0$.
\end{assumption}

We argue the above assumption is mild because input $X$ and self-supervised signal $S$ are two different views of the data, and are expected to contain task-relevant information $Y$.
In Proposition~\ref{prop:bayes_error_rate_mutual_info}, we make connections between the Bayes error rate and pre-training losses, which explains why the proposed pre-training losses are helpful for downstream tasks. Proof in Appendix~\ref{section:information_theory_self_supervised}. 

\begin{proposition} \label{prop:bayes_error_rate_mutual_info}
We can upper bound Bayes error rate by $P_e \leq 1-\exp(-H(Y) + I(Z_X;X) - I(Z_X;X|Y))$,
and reduce the upper bound of $P_e$ by
$(1)$ maximizing the mutual information $I(Z_X; X)$ between the learned representation $Z_X$ and input $X$, which can be achieved by minimizing temporal reconstruction loss $\mathcal{L}_\text{recon}(\bm{\Theta})$, and $(2)$ minimizing the task-irrelevant information between the learned representation $Z_X$ and input $X$, which can be achieved by minimizing our multi-view loss $\mathcal{L}_\text{view}(\bm{\Theta})$.
\end{proposition}

The Proposition~\ref{prop:bayes_error_rate_mutual_info} suggests that if we can create a different views $S$ of our input data $X$ such that both $X$ and $S$ contain the task-relevant information $Y$, then by jointly optimizing two pre-training losses can result in the representation $Z_X$ with a lower Bayes error rate $P_e$. 
% Our analysis is based on the information theory framework developed in~\cite{tsai2020self}, in which they show that 
% using contrastive loss between $Z_X$ and $S$ (i.e., maximizing $I(Z_X;S)$), predicting $S$ from $Z_X$ (i.e., minimizing $H(S|Z_X)$), and predicting $Z_X$ from $S$ (i.e., minimizing $H(Z_X|S)$) can result in a smaller Bayes error rate $P_e$.

\section{Experiments}\label{section:experiments}

We evaluate \our using dynamic graph link prediction, which has been widely used in~\cite{sankar2018dynamic,goyal2018dyngem} to compare its performance with a variety of static and dynamic graph representation learning baselines.
% Our results on five publicly available datasets indicate \our significantly outperforms other models. 
Results on node classification is deferred to Appendix~\ref{section:node_cls}.
% and the implementation can be found \href{https://anonymous.4open.science/r/DyGraphTransformer-E705}{here}.
% Dataset and code can be found at \href{https://anonymous.4open.science/r/DyFormer-CF12}{\textcolor{blue}{here}}.

% \subsection{Experiment configurations}

%%%%%%%%%%%%%%%%%%%

\paragraph{Datasets.}
% We select real-world datasets with dynamic graph defined as a set of temporal ordered snapshot graph.
The detailed data statistics are summarized in Table~\ref{table:dataset_stat}, where the dynamic graph is defined as a set of temporal ordered snapshot graph.
Following the procedure as described in~\cite{sankar2018dynamic}, the graph snapshots are created by splitting the data using suitable time windows such that each snapshot has an equitable number of interactions. In each snapshot, the edge weights are determined by the number of interactions.
% Notice that we only consider datasets that defined dynamic graph as a set of temporal ordered snapshot graph.
% , and leave the study on other dynamic graph (i.e., continous time-step graph) as future direction.

\begin{table}[h]
\centering
\vspace{-5mm}
\caption{Dataset statistics.} \label{table:dataset_stat}
% \vspace{-3mm}
\scalebox{0.75}{
\begin{tabular}{lcccccc}
\hline\hline
                            %  & \textbf{\href{https://nrvis.com/download/data/dynamic/ia-radoslaw-email.zip}{\textcolor{black}{RDS}}} 
                            %  & \textbf{\href{http://konect.cc/networks/opsahl-ucsocial/}{\textcolor{black}{UCI}}} 
                            %  & \textbf{\href{https://www.yelp.com/dataset}{\textcolor{black}{Yelp}}} 
                            %  & \textbf{\href{https://grouplens.org/datasets/movielens/10m/}{\textcolor{black}{ML-10M}}} 
                            %  & \textbf{\href{http://snap.stanford.edu/jodie/wikipedia.csv}{\textcolor{black}{Wikipedia}}} 
                            %  & \textbf{\href{http://snap.stanford.edu/jodie/reddit.csv}{\textcolor{black}{Reddit}}}
                            %  \\ \hline\hline
                            & \textbf{RDS} 
                            & \textbf{UCI} 
                            & \textbf{Yelp} 
                            & \textbf{ML-10M} 
                            & \begin{tabular}[c]{@{}l@{}}\textbf{SNAP-} \\ \textbf{Wikipedia} \end{tabular} 
                            & \begin{tabular}[c]{@{}l@{}}\textbf{SNAP-} \\ \textbf{Reddit}\end{tabular}\\ \hline\hline
\textbf{$|\mathcal{V}|$}      & $167$        & $1,809$      & $6,569$       & $20,537$        &  $9,227$           & $11,000$  \\ \hline
\textbf{$|\mathcal{E}|$}      & $1,521$      & $16,822$     & $95,361$      & $43,760$        &  $157,474$         & $672,447$               \\ \hline
T & $100$        & $13$         & $16$          & $13$            &  $11$              &   $11$         \\ \hline\hline
\end{tabular}}
\vspace{-1mm}
\end{table}

%%%%%%%%%%%%%%%%%%%

\paragraph{Link prediction.}
To compare \our with baselines, we follow the evaluation strategy in~\cite{goyal2018dyngem,zhou2018dynamic,sankar2018dynamic} by training a logistic regression classifier taking two node embeddings as input for dynamic graph link prediction.
Specifically, we learn the dynamic node representations on snapshot graphs $\{\mathcal{G}_1, \ldots, \mathcal{G}_T\}$ and evaluate \our by predicting links at $\mathcal{G}_{T+1}$.
For evaluation, we consider all links in $\mathcal{G}_{T+1}$ as positive examples and an equal number of sampled unconnected node pairs as negative examples.
We split $20\%$ of the edge examples for training the classifier, $20\%$ of examples for hyper-parameters tuning, and the rest $60\%$ of examples for model performance evaluation following the practice of existing studies (e.g.,~\cite{sankar2018dynamic}).
We evaluate the link prediction performance using Micro and Macro scores, where the Micro is calculated across the link instances from all the time-steps while the Macro is computed by averaging the AUC at each time-step.
During inference, all nodes in the testing set (from $60\%$ edge samples in $\mathcal{G}_{T+1}$) are selected as the target nodes.
To scale the inference of the testing sets of any sizes, we compute the full-attention by first splitting all self-attentions into multiple chunks then iteratively compute the self-attention in each chunk.
Since only a fixed number of self-attention is computed at each iteration, we significantly reduce \oure's inference memory consumption.
We also repeat all experiments three times with different random seeds. 

%%%%%%%%%%%%%%%%%%%

\subsection{Experiment results}

\begin{table}[t]
\centering
\caption{Comparing \our with baselines using \textit{Micro}- and \textit{Macro}-\textit{AUC} on real-world datasets.}\label{table:single_step_link_prediction}
% \vspace{-3mm}
\scalebox{0.61}{
\begin{tabular}{llcccc}
\hline\hline
\textbf{Method}                              & \textbf{AUC}    & \textbf{RDS}        & \textbf{UCI}        & \textbf{Yelp}        & \textbf{ML-10M} \\ \hline\hline
\multirow{2}{*}{\textsc{Node2Vec}}           & \text{Micro}   & $81.10\pm0.87$      & $81.41\pm0.60$      & $68.93\pm0.33$       & $90.50\pm0.83$  \\ % \cline{2-8} 
                                             & \text{Macro}   & $82.85\pm0.86$      & $81.39\pm0.76$      & $67.38\pm0.49$       & $89.48\pm0.62$  \\ \hline
%%%%%%%%%%%%%%%%%%%%%%%%%%%%%%%%%%%%%%
\multirow{2}{*}{\textsc{GraphSAGE}}          & \text{Micro}   & $85.49\pm0.96$      & $79.85\pm2.62$      & $62.36\pm1.01$       & $86.31\pm0.97$  \\ % \cline{2-8} 
                                             & \text{Macro}   & $86.64\pm0.89$      & $78.45\pm2.01$      & $58.36\pm0.91$       & $90.23\pm0.90$  \\ \hline\hline
%%%%%%%%%%%%%%%%%%%%%%%%%%%%%%%%%%%%%%
\multirow{2}{*}{\textsc{DynAERNN}}           & \text{Micro}   & $80.56\pm0.77$      & $79.29\pm1.90$      & $71.54\pm0.83$       & $87.01\pm0.88$  \\ % \cline{2-8} 
                                             & \text{Macro}   & $80.16\pm0.91$      & $83.81\pm1.25$      & $72.29\pm0.58$       & $89.04\pm0.67$  \\ \hline
%%%%%%%%%%%%%%%%%%%%%%%%%%%%%%%%%%%%%%
\multirow{2}{*}{\textsc{DynGEM}}             & \text{Micro}   & $79.29\pm1.01$      & $76.36\pm0.83$      & $69.43\pm1.09$       & $79.80\pm0.88$  \\ % \cline{2-8} 
                                             & \text{Macro}   & $81.94\pm1.97$      & $78.22\pm0.99$      & $69.93\pm0.78$       & $84.86\pm0.49$  \\ \hline\hline
%%%%%%%%%%%%%%%%%%%%%%%%%%%%%%%%%%%%%%
\multirow{2}{*}{\textsc{DySAT}}              & \text{Micro}   & $83.89\pm0.92$      & $83.10\pm0.99$      & $69.00\pm0.22$       & $88.91\pm0.87$  \\ % \cline{2-8} 
                                             & \text{Macro}   & $83.60\pm0.68$      & $86.32\pm1.46$      & $69.42\pm0.25$       & $90.63\pm0.91$  \\ \hline
%%%%%%%%%%%%%%%%%%%%%%%%%%%%%%%%%%%%%%
\multirow{2}{*}{\textsc{EvolveGCN}}          & \text{Micro}   & $85.35\pm0.87$      & $85.81\pm0.50$      & $68.99\pm0.67$       & $92.79\pm0.21$             \\ % \cline{2-8} 
                                             & \text{Macro}   & $86.53\pm0.76$      & $84.18\pm0.72$      & $69.41\pm0.26$       & $93.45\pm0.19$             \\ \hline
%%%%%%%%%%%%%%%%%%%%%%%%%%%%%%%%%%%%%%
\multirow{2}{*}{\textbf{\oure}}              & \text{Micro}   & $\bm{88.77\pm0.50}$ & $\bm{87.91\pm0.32}$ & $\bm{73.39\pm0.21}$  & $\bm{95.30\pm0.36}$ \\ % \cline{2-8} 
                                             & \text{Macro}   & $\bm{89.77\pm0.46}$ & $\bm{88.49\pm0.43}$ & $\bm{74.31\pm0.23}$  & $\bm{96.16\pm0.22}$ \\ \hline\hline
\end{tabular}
}
\vspace{-2mm}
\end{table}

\begin{table}[t]
\vspace{-1mm}
\centering
\caption{Comparison of \textit{Micro}- and \textit{Macro}-\textit{AUC} on real-world datasets restricted to new edges. }\label{table:single_step_link_prediction_new_edge}
% \vspace{-4mm}
\scalebox{0.6}{
\begin{tabular}{llcccc}
\hline\hline
\textbf{Method}                              & \textbf{AUC}    & \textbf{RDS}        & \textbf{UCI}        & \textbf{Yelp}        & \textbf{ML-10M} \\ \hline\hline
\multirow{2}{*}{\textsc{Node2Vec}}           & \text{Micro}   & $75.62\pm1.42$      & $75.31\pm0.83$      & $68.83\pm0.29$       & $88.92\pm0.79$  \\ % \cline{2-8} 
                                             & \text{Macro}   & $76.25\pm0.85$      & $75.82\pm0.96$      & $68.00\pm0.51$       & $88.01\pm0.50$  \\ \hline
%%%%%%%%%%%%%%%%%%%%%%%%%%%%%%%%%%%%%%
\multirow{2}{*}{\textsc{GraphSAGE}}          & \text{Micro}   & $80.21\pm0.87$      & $76.56\pm1.91$      & $61.97\pm1.00$       & $85.18\pm0.89$  \\ % \cline{2-8} 
                                             & \text{Macro}   & $79.99\pm0.78$      & $75.94\pm1.88$      & $58.49\pm0.89$       & $89.31\pm0.93$  \\ \hline\hline
%%%%%%%%%%%%%%%%%%%%%%%%%%%%%%%%%%%%%%
\multirow{2}{*}{\textsc{DynAERNN}}           & \text{Micro}   & $68.43\pm1.13$      & $77.39\pm2.10$      & $70.82\pm0.93$       & $86.89\pm0.75$  \\ % \cline{2-8} 
                                             & \text{Macro}   & $68.18\pm1.23$      & $81.82\pm1.71$      & $71.56\pm0.77$       & $89.45\pm0.53$  \\ \hline
%%%%%%%%%%%%%%%%%%%%%%%%%%%%%%%%%%%%%%
\multirow{2}{*}{\textsc{DynGEM}}             & \text{Micro}   & $72.43\pm1.62$      & $74.72\pm0.73$      & $69.23\pm1.76$       & $77.18\pm1.96$  \\ % \cline{2-8} 
                                             & \text{Macro}   & $74.49\pm2.21$      & $76.34\pm0.78$      & $70.67\pm1.32$       & $82.62\pm0.49$  \\ \hline\hline
%%%%%%%%%%%%%%%%%%%%%%%%%%%%%%%%%%%%%%
\multirow{2}{*}{\textsc{DySAT}}              & \text{Micro}   & $76.28\pm1.34$      & $81.18\pm1.09$      & $69.12\pm0.21$       & $88.21\pm0.64$  \\ % \cline{2-8} 
                                             & \text{Macro}   & $76.87\pm1.21$      & $83.43\pm1.57$      & $69.20\pm0.20$       & $88.98\pm0.87$  \\ \hline
%%%%%%%%%%%%%%%%%%%%%%%%%%%%%%%%%%%%%%
\multirow{2}{*}{\textsc{EvolveGCN}}          & \text{Micro}   & $78.36\pm0.91$      & $81.99\pm0.73$      & $68.73\pm0.64$       & $90.91\pm0.32$             \\ % \cline{2-8} 
                                             & \text{Macro}   & $79.18\pm1.01$      & $82.18\pm0.76$      & $68.63\pm0.30$       & $91.45\pm0.29$             \\ \hline
%%%%%%%%%%%%%%%%%%%%%%%%%%%%%%%%%%%%%%
\multirow{2}{*}{\oure}                       & \text{Micro}   &$\bm{82.78\pm0.56}$ & $\bm{85.78\pm0.99}$ & $\bm{73.32\pm0.22}$  & $\bm{93.01\pm0.23}$ \\ % \cline{2-8} 
                                             & \text{Macro}   & $\bm{82.89\pm0.52}$ & $\bm{86.21\pm0.56}$ & $\bm{73.88\pm0.22}$  & $\bm{93.56\pm0.21}$ \\ \hline\hline
\end{tabular}
}
\vspace{-2mm}
\end{table}
%%%%%%%%%%%%%%%%%%%

\begin{figure*}[t]
    \centering
    \includegraphics[width=0.80\textwidth, height=2cm]{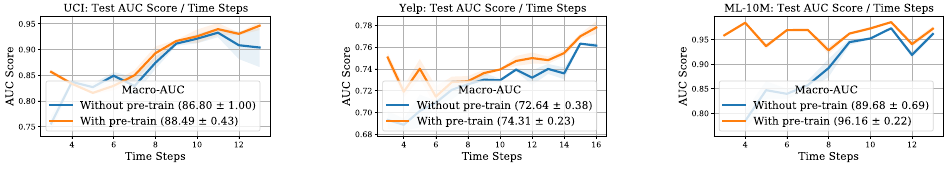}
    \vspace{-6mm}
    \caption{Comparison of the \textit{Micro}- and \textit{Macro}-\textit{AUC} score of \our with and without pre-training.}
    \label{fig:compare_with_without_pretrain} 
    \vspace{-3mm}
\end{figure*}

\begin{figure*}[t]
    \centering
    \includegraphics[width=.95\textwidth, height=2cm]{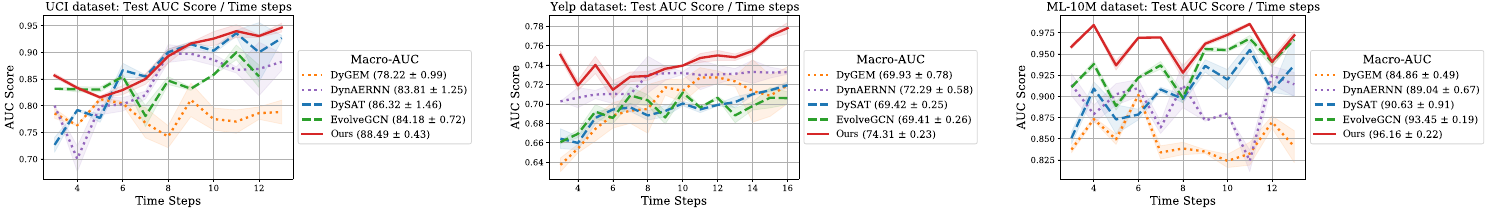}
    \vspace{-6mm}
    \caption{Comparison of per-step AUC-score and Macro AUC-score across multiple time steps.
    % , where the Macro score is reported in the box next to the curves.
    }
    \label{fig:time_step_curve}
    \vspace{-3mm}
\end{figure*}

\paragraph{The effectiveness of \oure.}
Table~\ref{table:single_step_link_prediction} indicates the state-of-the-art performance of our approach on link prediction tasks, where \our achieves a consistent $1\%\sim3\%$ Macro gain on all datasets.
Besides, \our is more stable when using different random seeds observed from a smaller standard deviation of the AUC score.
To better understand the behaviors of different methods from a finer granularity, we compare the model performance at each time-step in Figure~\ref{fig:time_step_curve}
and observe that the performance of \our is relatively more stable than other methods over time.
Besides, we additionally report the results of dynamic link prediction evaluated only on unseen links at each time-step.
Here, we define unseen links as the ones that first appear at the prediction time-step but are not in the previous graph snapshots.
From Table~\ref{table:single_step_link_prediction_new_edge},
we find that although all methods achieve a lower AUC score, which may be due to the new link prediction is more challenging, \our still achieves a consistent $1\%\sim 3\%$ Macro AUC-score gain.
% Moreover, we compare the training time and memory consumption with several baselines in Table~\ref{table:memory_time} 
% % (in Appendix~\ref{section:link_pred_appendix}) 
% and shows that \our maintains a good scalability. 
%########################################################
%########################################################
%########################################################

\paragraph{The effectiveness of pre-training.}
We compare the performance of \our with/without pre-training.
As shown in Figure~\ref{fig:compare_with_without_pretrain},  
% the Macro score of \our is similar to baselines without pre-training.~\yh{not necessary, we can remove this sentence} 
% However,
\our's performance is significantly improved if we first pre-train it with the self-supervised loss then fine-tuning on downstream tasks.
When comparing the AUC scores at each time-step, we observe that \our without pre-training has a lower performance but a larger variance. 
This may be due to the vast number of training parameters in \our, which potentially requires more data to be trained well.
The self-supervised pre-training alleviates this by utilizing additional unlabeled data. 

%########################################################
%########################################################
%########################################################
\paragraph{Results on dataset with missing/spurious links.}
We study the effect of noisy input on the performance of \our using \textit{UCI}\cite{kunegis2013konect,opsahl2009clustering} and \textit{Yelp} datasets. We achieve this by randomly selecting $10\%$, $20\%$, $50\%$ of the node pairs and changing their connection status either from connected to not-connected or from not-connected to connected. 
As shown in Table~\ref{eq:noisy_input}, although the performance of both using full-attention and 1-hop attention decreases as the noisy level increases, the performance of using full-attention aggregation is more stable and robust as the noisy level changes. 
This is because 1-hop attention relies more on the given structure, while full-attention only take the give structure as a reference and learns the ``ground truth'' graph structure by gradient descent update.

\begin{table}[t]
\vspace{-1mm}
\caption{Comparison of \our and its variants with input graph with different noisy level.} \label{eq:noisy_input}
% \vspace{-4mm}
\centering
\scalebox{0.68}{
\begin{tabular}{ l l l c c c }
\hline\hline
                       & \textbf{Method}                                       & \textbf{AUC}          & \textbf{10\%} & \textbf{20\%} & \textbf{50\%} \\ \hline\hline
\multirow{4}{*}{\textbf{UCI}}    & \multirow{2}{*}{1-hop attention}       & Micro & $82.97 \pm 0.56$ & $81.23 \pm 0.78$ & $77.85 \pm 0.66$ \\ % \cline{3-6} 
                                 &                                              & Macro & $83.01 \pm 0.61$ & $82.10 \pm 0.60$ & $78.43 \pm 0.67$ \\ \cline{2-6} 
                                 & \multirow{2}{*}{Full attention}      & Micro & $86.98 \pm 0.51$ & $86.10 \pm 0.57$ & $84.36 \pm 0.49$ \\ % \cline{3-6} 
                                 &                                              & Macro & $86.12 \pm 0.57$ & $85.93 \pm 0.59$ & $85.51 \pm 0.51$ \\ \hline
\multirow{4}{*}{\textbf{Yelp}}   & \multirow{2}{*}{1-hop attention}       & Micro & $70.00 \pm 0.20$ & $68.55 \pm 0.21$ & $65.32 \pm 0.22$ \\ % \cline{3-6} 
                                 &                                              & Macro & $69.94 \pm 0.20$ & $68.45 \pm 0.23$ & $65.61 \pm 0.15$ \\ \cline{2-6} 
                                 & \multirow{2}{*}{Full attention}      & Micro & $70.99 \pm 0.20$ & $71.74 \pm 0.19$ & $70.93 \pm 0.21$ \\ % \cline{3-6} 
                                 &                                              & Macro & $71.64 \pm 0.18$ & $71.67 \pm 0.21$ & $69.93 \pm 0.21$ \\ \hline\hline
\end{tabular}
}
\vspace{-1mm}
\end{table}

\paragraph{More results.}
Due to the space limit, more ablation study results are deferred to Appendix~\ref{section:more_ablation_study}, which includes ablation study on the effectiveness of spatial-temporal encoding, the number of layers in \our, two-tower and single-tower model architecture, self-attention mechanism, and computation cost.

\section{Conclusion}
In this paper, we introduce \our for dynamic graph representation learning, which can efficiently leverage the graph topology and capture implicit edge connections.
To further improve the generalization ability, two complementary pre-training tasks are introduced.
To handle large-scale dynamic graphs, a temporal-union graph structure and a target-context node sampling strategy are designed for an efficient and scalable training.
Extensive experiments on real-world dynamic graphs show that \our presents significant performance gains over several state-of-the-art baselines.
Potential future directions include exploring GNNs on continuous dynamic graphs and studying its expressive power.

\section*{Acknowledgements}
This work was supported in part by NSF grant 2008398.
% \clearpage

\bibliographystyle{plain}
\bibliography{reference}

\clearpage
\appendix

\section{More experiment results} \label{section:more_ablation_study}

\paragraph{The effectiveness of spatial-temporal encoding.}
In Table~\ref{table:compare_positional_encoding}, we conduct an ablation study by independently removing two encodings to validate the effectiveness of spatial-temporal encoding.
We observe that even without any encoding (i.e., ignoring the spatial-temporal graph topologies), due to full-attention, \our is still very competitive comparing with the state-of-the-art baselines in Table~\ref{table:single_step_link_prediction}. 
However, we also observe a $0.6\% \sim4.6\%$ performance gain when adding the spatial connection and temporal distance encoding, which empirically shows their effectiveness. 

\begin{table}[h]
\vspace{-1mm}
\centering
\caption{Comparison of the \textit{Micro}- and \textit{Macro}-\textit{AUC} of with and without temporal-connection (TC) and spatial-distance (SD) encoding  on the real-world datasets. 
% \weilin{update micro score}
}\label{table:compare_positional_encoding}
% \vspace{-4mm}
\scalebox{0.7}{
\begin{tabular}{llcccc}
\hline\hline
\textbf{Method}                    & \textbf{AUC}    &  \textbf{UCI}       & \textbf{Yelp}       & \textbf{ML-10M} \\ \hline\hline
\multirow{2}{*}{Both encoding}
                                  & \text{Micro}   & $\bm{87.91\pm0.32}$ & $\bm{73.39\pm0.21}$ & $\bm{95.30\pm0.36}$ \\ % \cline{2-8} 
                                  & \text{Macro}   & $\bm{88.49\pm0.43}$ & $\bm{74.31\pm0.23}$ & $\bm{96.16\pm0.22}$  \\ \hline
\multirow{2}{*}{Without encoding} 
                                  & \text{Micro}   & $83.27\pm0.29$      & $72.82\pm0.37$      & $91.81\pm0.43$  \\ % \cline{2-8} 
                                  & \text{Macro}   & $83.87\pm0.47$      & $73.80\pm0.38$      & $92.59\pm0.35$  \\ \hline
\multirow{2}{*}{Only TC encoding}
                                  & \text{Micro}   & $84.78\pm0.31$      & $73.36\pm0.26$      & $94.51\pm0.37$  \\ % \cline{2-8} 
                                  & \text{Macro}   & $84.60\pm0.42$      & $74.31\pm0.25$      & $95.43\pm0.29$  \\ \hline
\multirow{2}{*}{Only SD encoding}
                                  & \text{Micro}   & $87.01\pm0.46$      & $72.98\pm0.32$      & $92.34\pm0.40$  \\ % \cline{2-8} 
                                  & \text{Macro}   & $87.99\pm0.47$      & $73.90\pm0.36$      & $93.13\pm0.33$  \\ \hline
                                  \hline
\end{tabular}
}
\vspace{-1mm}
\end{table}

\paragraph{The effectiveness of stacking more layers.}
% In classical GNNs, each individual layer only considers the local interactions. 
% To model longer-range dependencies, GNNs require a deep architecture so that information can be propagated from distant nodes. 
When stacking more layers, traditional GNNs usually suffer from the over-smoothing~\cite{zhao2019pairnorm, yan2021two} and result in a degenerated performance.
We study the effect of applying more \our layers and show results in Table~\ref{table:number_of_layers}.
In contrast to previous studies, \our has a relatively stable performance and does not suffer much from performance degradation when the number of layers increases.
This is potentially due to that \our only requires a shallow architecture since each individual layer is capable of modeling longer-range dependencies due to full-attention. 
Besides, the self-attention mechanism can automatically attend importance neighbors, therefore alleviate the over-smoothing and bottleneck effect.

\begin{table}[h]
\vspace{-1mm}
\centering
\caption{Comparison of the \textit{Micro}- and \textit{Macro}-\textit{AUC} score of \our with different number of layers. } \label{table:number_of_layers}
% \vspace{-4mm}
\scalebox{0.7}{
\begin{tabular}{llccc}
\hline\hline
\textbf{Method}                & \textbf{AUC}    & \textbf{UCI}        & \textbf{Yelp}        & \textbf{ML-10M}     \\ \hline\hline

\multirow{2}{*}{$2$ layers}    & \text{Micro-AUC}   & $87.89\pm0.43$      & $ 74.30\pm0.21$      & $ 94.99\pm0.21$     \\ % \cline{2-8} 
                               & \text{Macro-AUC}   & $88.31\pm0.53$      & $ 74.29\pm0.23$      & $ 96.08\pm0.15$     \\ \hline
\multirow{2}{*}{$4$ layers}    & \text{Micro-AUC}   & $87.42\pm0.36$      & $ \bm{73.39\pm0.21} $     & $95.30\pm0.36$  \\ % \cline{2-8} 
                               & \text{Micro-AUC}   & $88.35\pm0.37$      & $ \bm{74.31\pm0.23}$ & $ \bm{96.16\pm0.22}$\\ \hline
\multirow{2}{*}{$6$ layers}    & \text{Micro-AUC}   & $\bm{87.91\pm0.32}$ & $ 74.30\pm0.20$      & $ \bm{95.35\pm0.28}$      \\ % \cline{2-8} 
                               & \text{Micro-AUC}   & $\bm{88.49\pm0.43}$ & $ 74.28\pm0.22$      & $ 96.11\pm0.18$     \\ \hline\hline
% \multirow{2}{*}{$8$ layers}    & \text{Micro}   &                     &                     &                     \\ % \cline{2-8} 
%                               & \text{Micro}   &                     &                     &                     \\ \hline\hline
\end{tabular}
}
\vspace{-1mm}
\end{table}

%%%%%%%%%%%%%%%%%%%%%%
\paragraph{Comparing two-tower to single-tower architecture.}
In  Table~\ref{table:single_two_tower}, we compare the performance of \our with single- and two-tower design where a single-tower means a full-attention of over all pairs of target and context nodes.
We observe that the two-tower \our has a consistent performance gain ($0.5\%$ Micro- and Macro) over the single-tower on Yelp and ML-10M~\cite{harper2015movielens}.
This may be due to that the nodes within the target or context node set are sampled independently while inter-group nodes are likely to be connected. 
Only attending inter-group nodes helps \our better capturing these contextual information without fusing representations from irrelevant nodes.

\begin{table}[t]
\centering
\caption[caption with footnote]{Comparison of the \textit{Micro}- and \textit{Macro}-\textit{AUC} of \our using single-tower and two-tower model architecture on the real-world datasets.}\label{table:single_two_tower}
% \vspace{-4mm}
\scalebox{0.75}{
\begin{tabular}{llcccc}
\hline\hline
\textbf{Method}                    & \textbf{AUC}    &  \textbf{UCI}       & \textbf{Yelp}       & \textbf{ML-10M} \\ \hline\hline
\multirow{2}{*}{Single-tower } 
                                   & \text{Micro}   & $87.86\pm0.60$     & $72.95\pm0.20$     & $94.80\pm0.81$  \\ % \cline{2-8} 
                                   & \text{Macro}   & $88.27\pm0.68$      & $73.81\pm0.21$      & $95.49\pm0.57$  \\ \hline
\multirow{2}{*}{Two-tower}
                                   & \text{Micro}   & $\bm{87.91\pm0.32}$ & $\bm{73.39\pm0.21}$ & $\bm{95.30\pm0.36}$ \\ % \cline{2-8} 
                                   & \text{Macro}   & $\bm{88.49\pm0.43}$ & $\bm{74.31\pm0.23}$ & $\bm{96.16\pm0.22}$  \\ \hline\hline
\end{tabular}
}
\vspace{-1mm}
\end{table}

\paragraph{Comparing $K$-hop attention with full-attention.}
To better understand full-attention, we compare it with $1$-hop and $3$-hop attention.
These variants are evaluated based on the single-tower~\our to include all node pairs into consideration. 
Table~\ref{table:compare_k_hop_vs_full_attention} shows the results where we observe that the full-attention presents a consistent performance gain around $1\%\sim 3\%$ over the other two variants.
This demonstrates the benefits of full-attention when modeling implicit edge connections in graphs with a larger receptive fields comparing to its $K$-hop counterparts.
\begin{table}[h]
\vspace{-1mm}
\centering
\caption{Comparison of the \textit{Micro}- and \textit{Macro}-\textit{AUC} of \textit{full attention} and \textit{$K$-hop attention}  using the single-tower architecture on the real-world datasets. }\label{table:compare_k_hop_vs_full_attention}
% \vspace{-4mm}
\scalebox{0.7}{
\begin{tabular}{llcccc}
\hline\hline
\textbf{Method}                    & \textbf{AUC}    &  \textbf{UCI}       & \textbf{Yelp}       & \textbf{ML-10M} \\ \hline\hline
\multirow{2}{*}{Full attention } 
                                  & \text{Micro}   & $\bm{87.86\pm0.60}$ & $\bm{72.95\pm0.20}$ & $\bm{94.80\pm0.81}$ \\ % \cline{2-8} 
                                  & \text{Macro}   & $\bm{88.27\pm0.68}$ & $\bm{73.81\pm0.21}$ & $\bm{95.49\pm0.57}$  \\ \hline
\multirow{2}{*}{$1$-hop neighbor} 
                                  & \text{Micro}   & $84.62\pm0.31$      & $71.33\pm0.43$      & $91.88\pm0.73$  \\ % \cline{2-8} 
                                  & \text{Macro}   & $85.10\pm0.15$      & $71.45\pm0.45$      & $92.18\pm0.44$  \\ \hline
\multirow{2}{*}{$3$-hop neighbor}  
                                  & \text{Micro}   & $87.01\pm0.89$     & $71.19\pm0.22$      & $91.83\pm0.92$    \\ % \cline{2-8} 
                                  & \text{Macro}   & $87.48\pm0.88$     & $72.31\pm0.22$      & $92.33\pm0.82$  \\ \hline\hline      
\end{tabular}
}
\vspace{-1mm}
\end{table}

%%%%%%%%%%%%%%%%%%%%%%%%%%%%%%%%%%%%%%%%%%%%

\paragraph{Computation time and memory consumption.}
In Table~\ref{table:memory_time}, we compare the memory consumption and epoch time on the last time step of ML-10M and Yelp dataset. 
We chose the last time step of these two datasets because its graph size is relatively larger than others, which can provide a more accurate time and memory estimation. 
The memory consumption is record by  \texttt{nvidia-smi} and the time is recorded by function \texttt{time.time()}. 
During pre-training, \our samples $256$ context node and $256$ context node at each iteration. During fine-tuning, \our first $256$ positive links (links in the graph) and sample $2,560$ negative links (node pairs that do not exist in the graph), then treat all nodes in the sampled node pairs at target nodes and sample the same amount of context nodes. Notice that although the same sampling size hyper-parameter is used, since the graph size and the graph density are different, the actual memory consumption and time are also different. 
For example, since the Yelp dataset has more edges with more associated nodes for evaluation than ML-10M, the memory consumption and time are required on Yelp than on ML-10M dataset.
\begin{table}[h]
\centering
\caption[caption with footnote]{Comparison of the epoch time and memory consumption of \our with baseline methods.
% on the last timestep of \textbf{ML-10M} and \textbf{Yelp} dataset.
}\label{table:memory_time}
\scalebox{0.7}{
\begin{tabular}{llll}
\hline\hline
\textbf{Dataset}    & \textbf{Method}        &  \textbf{Memory}       & \textbf{Epoch / Total time} \\ \hline\hline
\multirow{4}{*}{\textbf{ML-10M}} 
                    & \textsc{DySAT}         &  $9.2$GB                          & $97.2$s/$4276.8$s ($45$ epochs)    \\ 
                    & \textsc{EvolveGCN}     &  $13.6$GB                         & $6.9$s/$821.1$s ($120$ epochs)    \\ 
                    & \our (Pretrain)    &  $6.5$GB                         & $38.9$s/$986.5$s ($89$ epochs)               \\ 
                    & \our (Finetune)     &  $10.1$GB                         & $2.98$s/$62.2$s  ($22$ epochs)                 \\ \hline
\multirow{4}{*}{\textbf{Yelp}} 
                    & \textsc{DySAT}         &  $5.4$GB                          & $29.4$s/$4706.4$s ($160$ epochs)   \\ 
                    & \textsc{EvolveGCN}     &  $7.5$GB                          & $19.14$s/$1091.2$s ($57$ epochs)                 \\ 
                    & \our (Pretrain)     &  $21.3$GB                         & $11.8$s/$413.5$s ($34$ epochs)     \\ 
                    & \our (Finetune)     &  $21.3$GB                         & $21.41$s/$521.6$s ($23$ epochs)                 \\ \hline\hline
\end{tabular}
}
\vspace{-2mm}
\end{table}

% \begin{table}[h]
% \centering
% \scalebox{0.85}{
% \begin{tabular}{lcccc}
% \hline\hline
% \textbf{Method}        &  \textbf{Memory consumption}       & \textbf{Epoch training time}       & \textbf{Total training time} \\ \hline\hline
% \textsc{DySAT}         &  $9.2$GB                          & $97.2$s                         &  $4276.8$s                \\ \hline
% \textsc{EvolveGCN}     &  $13.6$GB                         & $6.9$ s                         &  $821.1$s                \\ \hline
% \our (PT)    &  $13.2$GB                         & $38.9$s                         &  $986.5$s                \\ \hline
% \our (FT)     &  $7.8$GB                          & $1.34$s                         &  $40.2$s                  \\ \hline\hline
% \end{tabular}
% }
% \vspace{-5pt}
% \caption[caption with footnote]{Comparison of the epoch training time and memory consumption of \our with baseline models on the last timestep of \textbf{Yelp-16} dataset.}\label{table:memory_time}
% \end{table}

\section{Node classification results} \label{section:node_cls}
\begin{figure}[t]
    \centering
    \includegraphics[width=0.49\textwidth]{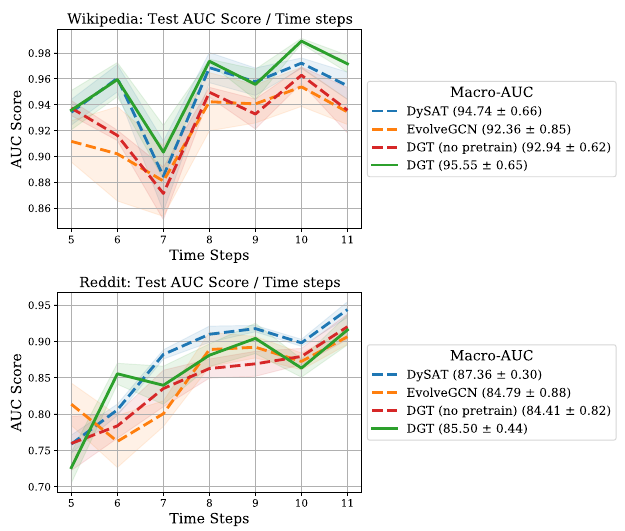}
    \vspace{-3mm}
    \caption{Comparison of \our with baselines across multiple time steps, where the Macro score is reported in the box next to the curves}
    \label{fig:node_cls}
\end{figure}

In this section, we show that although \our is orginally designed for the link prediction task, the learned representation of \our can be also applied to binary node classification.
We evaluate \our on SNAP-Wikipedia and SNAP-Reddit dataset~\cite{snapnets}, where dataset statistic is summarized in Table~\ref{table:dataset_stat}.
The snapshot is created in a similar manner as the link prediction task.
% Similar to the link prediction task, snapshot graphs are created by splitting the data using suitable time windows such that each snapshot has an equitable and reasonable number of interactions. In each snapshot, the edge weight is determined by the number of interactions between the associated node pairs during that time duration.
As shown in Table~\ref{table:node_cls} and Figure~\ref{fig:node_cls}, \our performs around $0.7\%$ better than all baselines on the SNAP-Wikipedia dataset and around $0.7\%$ better than \textsc{EvolveGCN} on SNAP-Reddit dataset. However, the results \our on the SNAP-Reddit dataset\footnote{The Reddit dataset are collected and released by SNAP at Stanford University. Please refer to~\url{http://snap.stanford.edu/jodie/} for details.} is slightly lower than \textsc{DySAT}. This is potentially due to \our is less in favor of a dense graph, e.g., SNAP-Reddit dataset, with very dense graph structure information encoded by spatial-temporal encodings.

\begin{table}[h]
\centering
\caption{Comparison of the \textit{Micro}- and \textit{Macro}-\textit{AUC} on the real-world datasets for binary node classification task. } \label{table:node_cls}
% \vspace{-3mm}
\scalebox{0.72}{
\begin{tabular}{llcc}
\hline\hline
\textbf{Method}                & \textbf{AUC}    
                               & \begin{tabular}[c]{@{}l@{}} \textbf{SNAP-} \\ \textbf{Wikipedia}\end{tabular} 
                               & \begin{tabular}[c]{@{}l@{}} \textbf{SNAP-} \\ \textbf{Reddit}\end{tabular}     \\ \hline\hline
% \multirow{2}{*}{\textsc{GAT+LSTM}}    
%                               & \text{Micro}   &                     &                     \\ % \cline{2-8} 
%                               & \text{Micro}   & $94.04\pm0.72$      & $81.86\pm2.19$      \\ \hline\hline
\multirow{2}{*}{\textsc{DySAT}}
                               & \text{Micro}   & $94.69\pm0.46$      & $\bm{87.35\pm0.28}$      \\ % \cline{2-8} 
                               & \text{Macro}   & $94.74\pm0.66$      & $\bm{87.36\pm0.30}$        \\ \hline
\multirow{2}{*}{\textsc{EvolveGCN}}    
                               & \text{Micro}   & $92.31\pm0.68$      & $84.72\pm0.89$        \\ % \cline{2-8} 
                               & \text{Macro}   & $92.36\pm0.85$      & $84.79\pm0.88$      \\ \hline
\multirow{2}{*}{\our (w/o pre-training)}    
                               & \text{Micro}   & $92.90\pm0.84$      & $ 82.37\pm0.78$            \\ % \cline{2-8} 
                               & \text{Macro}   & $92.94\pm0.62$      & $ 84.41\pm0.82$      \\ \hline
\multirow{2}{*}{\our (w/ pre-training)}    
                               & \text{Micro}   & $\bm{95.49\pm0.66}$ & $85.48\pm0.43$      \\ % \cline{2-8} 
                               & \text{Macro}   & $\bm{95.55\pm0.65}$ & $85.50\pm0.44$      \\ \hline\hline
\end{tabular}
}
\end{table}

\section{Experiment configuration} \label{section:exp_config}
\subsection{Hardware specification and environment}\label{section:hardware}
We run our experiments on a single machine with Intel i$9$-$10850$K, Nvidia RTX $3090$ GPU, and 32GB RAM memory. 
The code is written in Python $3.7$ and we use PyTorch $1.4$ on CUDA $10.1$ to train the model on the GPU. 

\subsection{Baseline hypter-parameters tuning}\label{section:hyper_parameters}

\paragraph{Baselines.}\footnote{We only compare with dynamic graph algorithms that takes a set of temporal ordered snapshot graph as input, and leave the study on other dynamic graph structure (e.g., continuous time-step algorithms~\cite{kumar2019predicting,xu2020inductive,rossi2020temporal} and datasets) as a future direction.}
We compare with several state-of-the-art methods as baselines including both static and dynamic graph learning algorithms.
For static graph learning algorithms, we compare against \textsc{Node2Vec}\cite{grover2016node2vec} and \textsc{GraphSAGE}\cite{hamilton2017inductive}.
To make the comparison fair, we feed these static graph algorithms the same temporal-union graph used in \our rather than any single graph snapshots.
For dynamic graph learning algorithms, we compare against \textsc{DynAERNN}\cite{goyal2020dyngraph2vec}, \textsc{DynGEM}\cite{goyal2018dyngem}, \textsc{DySAT}\cite{sankar2018dynamic}, and EvolveGCN~\cite{pareja2020evolvegcn}. 
We use the official implementations for all baselines and select the best hyper-parameters for both baselines and \oure.
% Notice that \emph{we only compare with dynamic graph algorithms that takes a set of temporal ordered snapshot graph as input, and leave the study on other dynamic graph structure (e.g., continuous time-step algorithms~\cite{kumar2019predicting,xu2020inductive,rossi2020temporal} and datasets) as a future direction.}

\paragraph{Model setups.}
The hyper-parameters for each dataset is selected using grid search on the validation set. 
More specifically, we select the negative sampling ratio (i.e., number of positive edge/number of negative edge) in the range $\{0.01, 0.1, 1\}$, number of the self-attention head is selected in the range $\{8, 16\}$, feature dimension is selected in the range $\{128, 256\}$, number of layers in the range $\{2, 4, 6\}$, maximum shortest path distance $D_{\max}$ in the range $\{2, 3, 5\}$ on the validation set, mini-batch size as $512$, and the pre-training loss weight $\gamma=1$. 

We tune the hyper-parameters of baselines following their recommended guidelines.

\textsc{Node2Vec}\footnote{\url{https://github.com/aditya-grover/node2vec}}: We use the default setting as introduced in~\cite{grover2016node2vec}. More specifically, for each node we use $10$ random walks of length $80$, context window size as $10$. The in-out hyper-parameter $p$ and return hyper-parameter $q$ are selected by grid-search in range $\{0.25, 0.5, 1, 2, 5\}$ on the validation set.

\textsc{GraphSAGE}\footnote{\url{https://github.com/williamleif/GraphSAGE}}: We use the default setting in~\cite{hamilton2017inductive}. More specifically, we train two layer GNN with neighbor sampling size $25$ and $10$. The neighbor aggregation is selected by grid-search from ``mean-based aggregation'', ``LSTM-based aggregation'', ``max-pooling aggregation'', and ``GCN-based aggregation'' on the validation set. 
In practice, GCN aggregator performs best on RDS~\cite{rossi2015network}, and UCI, and  max-pooling aggregator performs best on Yelp and ML-10M.

\textsc{DynGEM} and \textsc{DynAERNN}\footnote{\url{https://github.com/palash1992/DynamicGEM}}: We use the default setting as introduced in~\cite{goyal2018dyngem} and~\cite{goyal2020dyngraph2vec}. The scaling and regularization hyper-parameters is selected by grid-search in range $\alpha \in \{ 10^{-6}, 10^{-5}\}$, $\beta \in \{0.1, 1, 2, 5\}$, and $\nu_1,\nu_2 \in \{10^{-6}, 10^{-4}\}$ on the validation set.

\textsc{DySAT}\footnote{\url{https://github.com/aravindsankar28/DySAT}}: We use the default setting and model architecture as introduced in~\cite{sankar2018dynamic}. The co-occurring positive node pairs are sampled by running 10 random walks of length 40 for each node. The negative sampling ratio is selected by grid-search in the range $\{0.01, 0.1, 1\}$, number of the self-attention head is selected in the range $\{8, 16\}$, and the feature dimension is selected in the range $\{128, 256\}$ on the validation set.

\textsc{EvolveGCN}\footnote{\url{https://github.com/IBM/EvolveGCN}}: We use the default setting and model architecture as introduced in~\cite{pareja2020evolvegcn}. We train both EvolveGCN-O and EvolveGCN-H and report the architecture with the best performance on the validation set. In practice, EvolveGCN-O performs best on UCI, Yelp, and ML-10M, EvolveGCN-H performs best on Enron and RDS.

\section{Pre-training can reduce the irreducible error} \label{section:information_theory_self_supervised}

\subsection{Preliminary}

\noindent\textbf{Data processing inequality.}~
Random variables $X,Y,Z$ are said to form a \textit{Markov chain} $X\rightarrow Y \rightarrow Z$ if the joint probability mass function can be written as $P(x,y,z) = p(x) p(y|x) p(z|y)$.
 Suppose random variable $X,Y,Z$ forms a Markov chain $X\rightarrow Y \rightarrow Z$,
%  i.e., $X$ and $Y$ are conditional independent given $Z$,
 then we have $I(X;Y) \geq I(X;Z)$.

\noindent\textbf{Bayes error and entropy.}~
In the binary classification setting, \textit{Bayes error rate} is the lowest possible test error rate (i.e., irreducible error), which can be formally defined as
\begin{equation}
    P_e = \mathbb{E} \left[1 - \max_{y} p(Y=y|X) \right],
\end{equation}
where $Y$ denotes label and $X$ denotes input.
\cite{feder1994relations} derives an upper bound showing the relation between \textit{Bayes error rate} with entropy: 
\begin{equation} \label{eq:bayes_error_rate}
    - \log(1-P_e) \leq H(Y|X).
\end{equation}
The above inequality is used as the foundation of our following analysis.

\subsection{Proof of Proposition~\ref{prop:bayes_error_rate_mutual_info}}

In the following, we utilize the analysis framework developed in~\cite{tsai2020self} to show the importance of two pre-training loss functions.
By using Eq.~\ref{eq:bayes_error_rate}, we have $- \log(1-P_e) \leq H(Y|Z_X)$.
By rearanging the above inequality, we have the following upper bound on the Bayes error rate
\begin{equation}\label{eq:bayes_error_rate_rearange}
    \begin{aligned}
    P_e 
    &\leq 1- \frac{1}{\exp\big(H(Y|Z_X)\big) } \\
    % \exp\Big( - \big( H(Y|Z_X)\big) \Big) \\
    &\underset{(a)}{=} 1-\frac{1}{\exp\big( H(Y) - I(Z_X;Y) \big)} \\
    &\underset{(b)}{=} 1-\frac{1}{\exp\big( H(Y) - I(Z_X;X) + I(Z_X;X|Y) \big)},
    \end{aligned}
\end{equation}
where equality $(a)$ is due to $I(Z_X;Y) = H(Y) - H(Y|Z_X) $,
equality $(b)$ is due to $I(Z_X;Y) = I(Z_X;X) - I(Z_X;X|Y) + I(Z_X;Y|X)$ and $I(Z_X;Y|X)=0$ because $Z_X=f(X)$ is a deterministic mapping given input $X$.
Our goal is to find the deterministic mapping function $f$ to generate $Z_X$ that can maximize $I(Z_X;X) - I(Z_X;X|Y)$, such that the upper bound on the right hand side of Eq.~\ref{eq:bayes_error_rate_rearange} is minimized. We can achieve this by:
\begin{itemize} [noitemsep,topsep=0pt,leftmargin=3mm ]
    \item Maximizing the mutual information $I(Z_X;X)$ between the representation $Z_X$ to the input $X$.
    \item Minimizing the task-irrelevant information $I(Z_X;X|Y)$, i.e., the mutual information between the representation $Z_X$ to the input $X$ given task-relevant information $Y$.
\end{itemize}

In the following, we first show that minimizing $\mathcal{L}_\text{recon}(\mathbf{\Theta})$ can maximize the mutual information $I(Z_X;X)$, then we show that minimizing $\mathcal{L}_\text{view}(\mathbf{\Theta})$ can minimize the task irrelevant information $I(Z_X;X|Y)$.

\noindent\textbf{Maximize mutual information $I(Z_X;X)$.}~
By the relation between mutual information and entropy $I(Z_X;X) = H(X) - H(X|Z_X)$,
% . since $H(X)$ only dependent on the raw feature and is irrelevant to feature representation $Z_X$, 
we know that maximizing the mutual information $I(Z_X;X)$ is equivalent to minimizing the conditional entropy $ H(X|Z_X)$.
Notice that we ignore $H(X)$ because it is only dependent on the raw feature and is irrelevant to feature representation $Z_X$.
By the definition of conditional entropy, we have
\begin{equation}\label{eq:condition_entropy_log_likelihood}
    \begin{aligned}
    H(X|Z_X) &= \sum_{z_x\in\mathcal{Z}_\mathcal{X}} p(z_x) H(X|Z_X=z_x) \\
    &= \sum_{z_x\in\mathcal{Z}_\mathcal{X}} p(z_x) \sum_{x\in\mathcal{X}} - p(x|z_x) \log p(x|z_x)\\
    &= \sum_{z_x\in\mathcal{Z}_\mathcal{X}} \sum_{x\in\mathcal{X}} - p(x,z_x) \log p(x|z_x) \\
    &= \mathbb{E}_{\mathrm{P}(X, Z_X)}\Big[- \log \mathrm{P}(X|Z_X)\Big] \\
    % &= \sum_{x \in \mathcal{X}} p(x) H(X=x | Z_X=z_x),~\text{where}~ z_x = f(x), \\
    % &= \sum_{x \in \mathcal{X}} p(x) - \log P(X=x | Z_X=z_x),~\text{where}~ z_x = f(x),
    &= \min_{Q_{\bm{\theta}}}~\mathbb{E}_{\mathrm{P}(X, Z_X)}\Big[- \log Q_{\bm{\theta}}(X|Z_X)\Big] \\
    &\quad - \mathrm{KL}\Big(\mathrm{P}(X|Z_X) \Vert Q_{\bm{\theta}}(X|Z_X)\Big) \\
    &\leq \min_{Q_{\bm{\theta}}}~\mathbb{E}_{\mathrm{P}(X, Z_X)}\Big[- \log Q_{\bm{\theta}}(X|Z_X)\Big]
    \end{aligned}
\end{equation}
where $Q_{\bm{\theta}}(\cdot | \cdot)$ is a variational distribution with $\bm{\theta}$ represent the parameters in $Q_{\bm{\theta}}$ and $\mathrm{KL}$ denotes KL-divergence.

Therefore, maximizing mutual information $I(Z_X;X)$ can be achieved by minimizing $\mathbb{E}_{\mathrm{P}_{X, Z_X}}[- \log Q_{\bm{\theta}}(X|Z_X)]$. By assuming $Q_{\bm{\theta}}$ as the categorical distribution and $\bm{\theta}$ as a neural network, minimizing $\mathbb{E}_{\mathrm{P}_{X, Z_X}}[- \log Q_{\bm{\theta}}(X|Z_X)]$ can be think of as introducing a neural network parameterized by $\bm{\theta}$ to predict the input $X$ from the learned representation $Z_X$ by minimizing the binary cross entropy loss.

% Notice that minimizing $\mathbb{E}_{\mathrm{P}_{X, Z_X}}[- \log \mathrm{Q}_{\bm{\Theta}}(X|Z_X)]$ is a lower bound of $\mathbb{E}_{\mathrm{P}_{X, Z_X}}[- \log \mathrm{P}(X|Z_X)]$.
% By assumption $Q$ is a categorical distribution, minimizing $\mathbb{E}_{\mathrm{P}_{X, Z_X}}[- \log \mathrm{Q}_{\bm{\Theta}}(X|Z_X)]$ 
% An alternative way is to minimizing $\mathbb{E}_{\mathrm{P}_{X, Z_X}}[- \log \mathrm{Q}_{\bm{\theta}}(X|Z_X)]$, which is a lower bound of $\mathbb{E}_{\mathrm{P}_{X, Z_X}}[- \log \mathrm{P}(X|Z_X)]$.
% Here we assume  using cross entropy loss and $\phi$ is a linear model $\text{Linear}(\cdot)$.

\noindent\textbf{Minimize the task irrelevant information $I(Z_X;X|Y)$.}~
Recall that in our setting, input $X$ is the node features of $\{\mathcal{V}_\text{target}, \mathcal{V}_\text{context}\}$ and the subgraph induced by $\{\mathcal{V}_\text{target}, \mathcal{V}_\text{context}\}$. The self-supervised signal $S$ is node features of $\{\mathcal{V}_\text{target}, \widetilde{\mathcal{V}}_\text{context}\}$ and the subgraph induced by $\{\mathcal{V}_\text{target}, \widetilde{\mathcal{V}}_\text{context}\}$. 
Therefore, it is natural to
% Before proceeding, let us 
make the following mild assumption on the input random variable $X$, self-supervised signal $S$, and task relevant information $Y$.
\begin{assumption}\label{assumption:X_S_T_assump}
We assume tall task-relevant information is shared between the input random variable $X$, self-supervised signal $S$, i.e., we have $I(X;Y|S) = 0$ and $I(S;Y|X)=0$.
\end{assumption}

In the following, we show that minimizing $I(Z_X;X|Y)$ can be achieved by minimizing $H(Z_X | S)$.
From data processing inequality, we have $I(X; Y | S) \geq I(Z_X; Y | S) \geq 0$. From Assumption~\ref{assumption:X_S_T_assump}, we have $I(X; Y | S) = 0$, therefore we know $I(Z_X; Y | S) = 0$.
By the relation between mutual information and entropy, we have
\begin{equation}
    \begin{aligned}
    I(Z_X;X|Y) &= H(Z_X | Y) - H(Z_X | X, Y) \\
    &\underset{(a)}{=} H(Z_X | Y) \\
    &= H(Z_X|S,Y) + I(Z_X;S|Y) \\
    &= H(Z_X | S) - I(Z_X; Y | S) + I(Z_X;S|Y) \\
    &\underset{(b)}{=} H(Z_X | S) + I(Z_X;S|Y) \\
    &\underset{(c)}{\leq} H(Z_X | S) + I(X;S|Y),
    \end{aligned}
\end{equation}
where equality $(a)$ is due to $H(Z_X | X, Y)=0$ since $Z_X=f(X)$ and $f$ is a deterministic mapping, equality $(b)$ is due to $I(Z_X, Y | S) = 0$, and inequality $(c)$ is due to data processing inequality.

From Eq.~\ref{eq:condition_entropy_log_likelihood}, we know that 
\begin{equation}
    \begin{aligned}
    H(Z_X|S) &= \mathbb{E}_{\mathrm{P}(S,Z_X)}[-\log \mathrm{P}(Z_X|S)] \\
    &\leq \min_{Q^\prime_{\bm{\phi}}}~\mathbb{E}_{\mathrm{P}(S, Z_X)}\Big[- \log Q^\prime_{\bm{\phi}}(Z_X|S)\Big].
    \end{aligned}
\end{equation}

By assuming $Q^\prime_{\bm{\phi}}$ as the Gaussian distribution and $\bm{\phi}$ as a neural network, minimizing $\mathbb{E}_{\mathrm{P}_{S, Z_X}}[- \log Q_{\bm{\phi}}(Z_X|S)]$ can be think of as introducing a neural network parameterized by $\bm{\phi}$ that take $S$ as input and output $Z_S = \text{NeuralNetwork}_{\bm{\phi}}(S)$, then minimize the mean-square error between $Z_X$  and $Z_S$.

% which leads to our second loss, i.e., predicting the representation $Z_X$ by using self-supervised signal $S$, where $S$ is an another view of the original input $X$.
% On the other hand, $I(X; S|T)$ is the amount of the shared information between the raw feature and the self-supervised signal excluding the task-relevant information. Hence,I(X; S|T) would be large if the downstream tasks requires only a portion of the shared information

% \weilin{We might also need to discuss at which condition the loss is cross entropy loss, and at which condition it is mean-square error loss. }

\end{document}